\PassOptionsToPackage{table,xcdraw,dvipsnames}{xcolor}
\documentclass[]{fairmeta}

\usepackage{xcolor}
\usepackage{wrapfig}
\usepackage{mathpazo}
\usepackage{tgpagella}
\usepackage{bbding}
\usepackage{graphicx}
\usepackage{pgfplots}
\usepackage{makecell}
\usepackage{enumitem}
\usepackage{mathtools}
\usepackage{xfrac}
\usepackage{nicefrac}
\usepackage{amssymb}
\usepackage{array}
\usepackage{tabularx}
\usepackage{lipsum}
\usepackage{algorithm}
\usepackage{algpseudocode}
\usepackage{xspace}
\usepackage{booktabs}
\usepackage{multirow}
\usepackage{pifont}
\usepackage{float}
\usepackage{url}
\usepackage{tikz}

\pgfplotsset{compat=1.18}

\newcommand{\modelname}{\texttt{WM-VLM}\xspace}
\newcommand{\lanterntwo}{Tetris-2D\xspace}
\newcommand{\lanternthree}{Tetris-3D\xspace}

\newcommand{\thinkmorphnavorigin}{ThinkMorph-SN\xspace}
\titleformat*{\paragraph}{\sffamily\bfseries}

\newcolumntype{L}[1]{>{\raggedright\let\newline\\\arraybackslash\hspace{0pt}}m{#1}}
\newcolumntype{C}[1]{>{\centering\let\newline\\\arraybackslash\hspace{0pt}}m{#1}}
\newcolumntype{R}[1]{>{\raggedleft\let\newline\\\arraybackslash\hspace{0pt}}m{#1}}
\newcolumntype{Y}{>{\centering\arraybackslash}X}

\newcommand{\firstpagebranding}{%
  \begin{tikzpicture}[remember picture,overlay]
    \node[anchor=north west,inner sep=0pt]
      at ([xshift=2.5cm,yshift=-0.55cm]current page.north west)
      {\includegraphics[height=0.55cm]{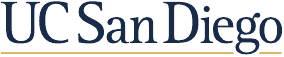}};
    \node[anchor=north west,inner sep=0pt]
      at ([xshift=5.75cm,yshift=-0.55cm]current page.north west)
      {\includegraphics[height=0.55cm]{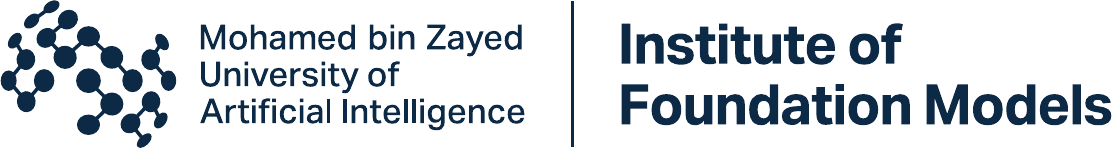}};
    \node[anchor=north west,inner sep=0pt]
      at ([xshift=10.35cm,yshift=-0.64cm]current page.north west)
      {\includegraphics[width=4cm,trim=348bp 898bp 360bp 900bp,clip]{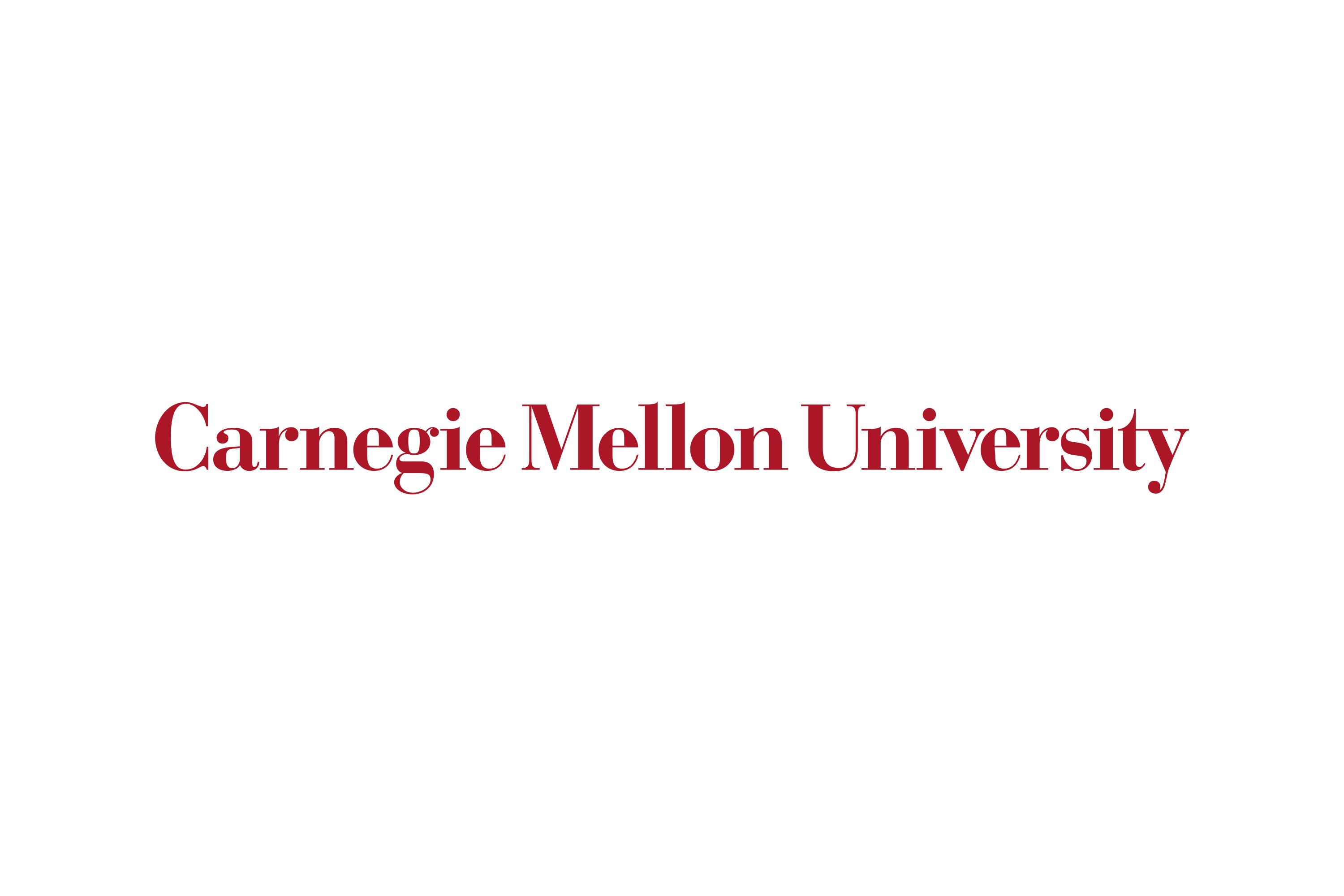}};
    \draw[metafg,line width=0.5pt]
      ([xshift=2.5cm,yshift=-1.55cm]current page.north west) --
      ([xshift=-2.5cm,yshift=-1.55cm]current page.north east);
  \end{tikzpicture}%
}

\title{WM-VLM: Probing Internal World Models for Interleaved Visual-Textual Reasoning}

\author[1,2,\dagger]{Yuheng Zha}
\author[1]{Yilei Wang}
\author[1]{Qiyue Gao}
\author[1]{Junrong Chen}
\author[1]{Yujia Wu}
\author[2]{Zhengfeng Lai}
\author[2]{Zhengzhong Liu}
\author[2,3]{Eric P. Xing}

\affiliation[1]{UC San Diego}
\affiliation[2]{Institute of Foundation Models, MBZUAI}
\affiliation[3]{Carnegie Mellon University}

\contribution[\dagger]{Work done while interning at IFM}

\abstract{
Humans often solve spatial problems by mentally simulating visual transformations. In contrast, conventional vision-language models (VLMs) reason primarily through language. We investigate whether VLMs can solve spatial problems by reasoning with both text and generated visual states. To this end, we introduce \modelname, which equips a pretrained VLM with a lightweight world model branch for generating intermediate visual states. Our two-stage training first teaches the model to generate the next visual state and then to use that state for reasoning. We programmatically construct spatial reasoning tasks with verifiable intermediate visual states. These tasks allow us to evaluate how well the model generates visual states and how much it relies on them to answer the question. On 2D and 3D mental rotation tasks, \modelname consistently outperforms the supervised fine-tuned backbone, with gains of up to 39.25 percentage points. Ablations suggest that these gains depend on the generated visual states, as removing or corrupting them sharply reduces performance. Together, these results suggest that internal world models offer a promising path toward more capable multimodal reasoning by enabling VLMs to reason jointly in language and visual space.
}

\date{\today}
\correspondence{Yuheng Zha <\email{yzha@ucsd.edu}>}
\metadata[Code]{\url{https://github.com/yuh-zha/WM-VLM}}
\metadata[Data and Checkpoints]{\url{https://huggingface.co/collections/yzha/wm-vlm}}

\begin{document}

\AddToHookNext{shipout/foreground}{\firstpagebranding}
\maketitle

\section{Introduction}
Spatial reasoning is a fundamental human ability that often involves mentally simulating visual transformations \citep{kosslyn1978visual,shepard1971mental}. For example, to determine whether turning right would bring them closer to a chair, people can mentally simulate the turn and its visual outcome. However, prior work has documented persistent weaknesses in the spatial reasoning capabilities of VLMs \citep{DBLP:conf/emnlp/KamathHC23a,DBLP:journals/corr/abs-2503-19707,zhang2026spinbench,DBLP:conf/cvpr/YangYGH0X25,DBLP:journals/corr/abs-2409-12969}. One possible limitation is that conventional VLMs perform intermediate reasoning primarily through text, even for visually grounded problems. Although recent reasoning VLMs perform well on mathematical and chart-based tasks \citep{huang2026visionr1,r1-onevision2025,chart-r1,zha2026vision,DBLP:journals/corr/abs-2508-09804}, many of these tasks can be solved primarily through language. Reasoning through text alone may fail to preserve the visual information needed to solve spatial problems.

\begin{figure}[htbp]
    \centering
    \includegraphics[width=0.48\linewidth]{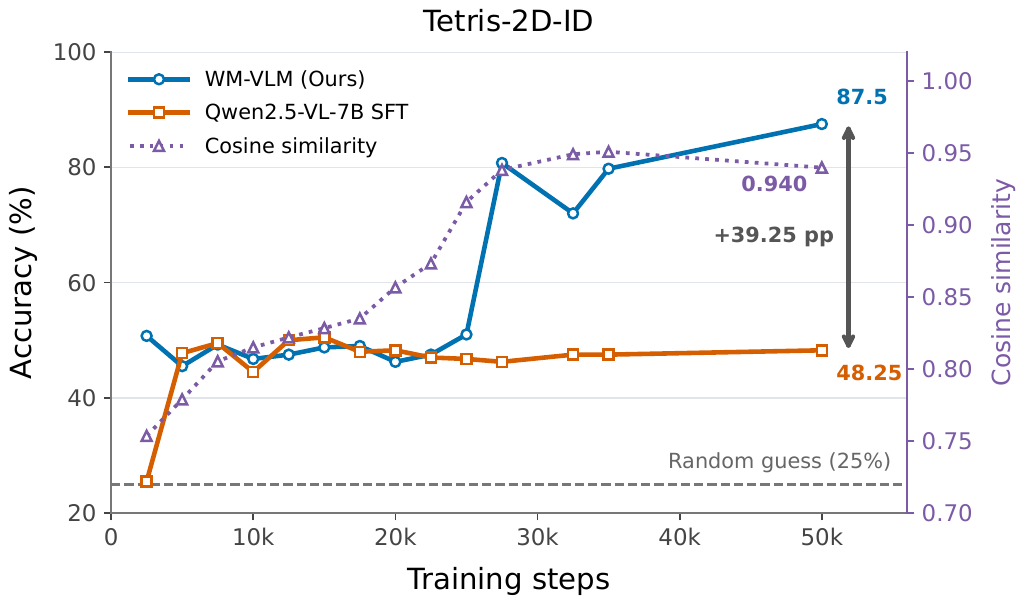}%
    \hfill
    \includegraphics[width=0.48\linewidth]{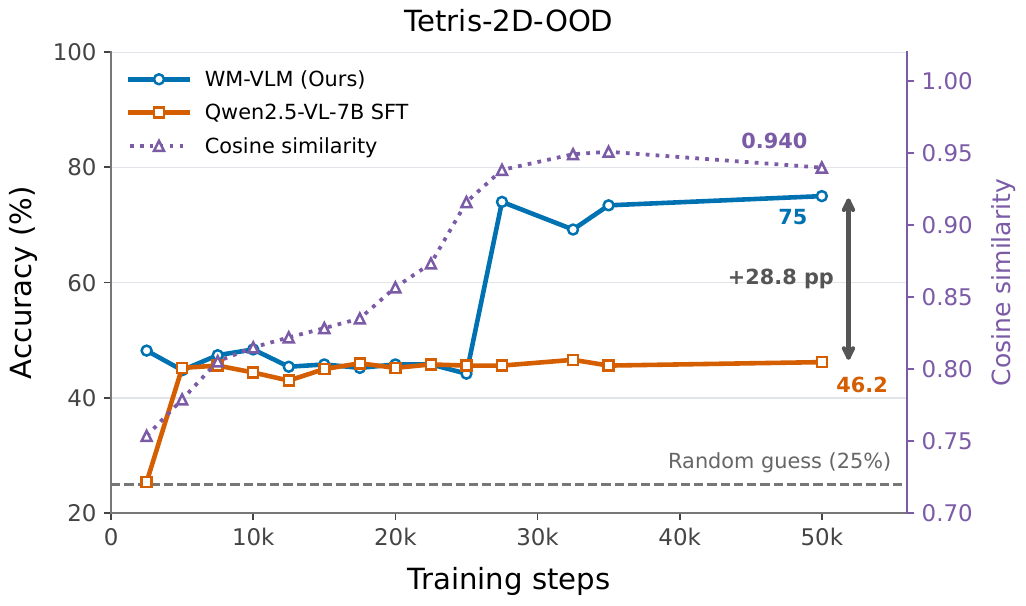}%
    \caption{The internal world model module and training objective positively contributes to solving spatial-related problems, which finally helps outperform the finetuned base VLM. \modelname exhibits a sudden performance transition at around step 25k, then plateaued, yielding a 39.25 points gain on ID set and 28.8 points gain on OOD set. We train \modelname with the middle-layer configuration and evaluate each checkpoint on \lanterntwo-ID (left) and \lanterntwo-OOD (right) splits. Training step indicates the training steps taken in Stage 1. All data points are obtained following an additional Stage 2 training. Each panel compares the accuracy of \modelname and Qwen2.5-VL-7B SFT (left axis). We also show the cosine similarity between the generated visual states and ground truth embedding during training (right axis, in purple). The dashed horizontal line indicates random-guess accuracy (25\%). }
    \label{fig:tetris-id-ood-accuracy-cosine}
   \end{figure}

Recent work seeks to enable VLMs to reason in both visual and textual spaces. Some methods use cropped or highlighted images as intermediate visual cues \citep{li2026lvr,liu2026dmlr,li2026livr,zhang2026unsilencing,gao2025icot}. These methods manipulate existing visual inputs rather than predict new visual states. Other methods interleave textual reasoning with generated visual states \citep{gu2026thinkmorph,mvot2025,mirage2026,wang2026monet,hu2026braid}, but they are often evaluated on planning tasks, where state prediction and action planning jointly determine performance. A separate line of work uses VLMs as reasoning policies that query external world models to generate imagined observations \citep{yang2026mindjourney,yu2026avic,zhu2026astra}. Although effective, these approaches delegate visual state prediction to a separate model, leaving open whether a VLM itself can learn to predict intermediate visual states that improve its spatial reasoning.

Motivated by work that frames visual generation as a form of world modeling \citep{wu2026visual,jin2026latentum}, we study whether an internal world model can improve spatial reasoning in VLMs. Standard VLM training typically supervises only text outputs, providing no direct objective for predicting intermediate visual states.
To address this limitation, we introduce \modelname, a VLM equipped with an internal world model. Inspired by Mixture-of-Transformers \citep{liang2024mixture}, \modelname incorporates a shallow generation branch that supports its internal world modeling capability. Accurately generating visual states, however, does not necessarily mean that the model can use them for reasoning. We therefore adopt a two-stage training strategy: the model first learns to generate intermediate visual states and then learns to reason with its own predictions. We train the model with both vision-language understanding and visual-state generation objectives. To isolate world modeling from action planning, we programmatically construct two controlled diagnostic datasets, \lanterntwo and \lanternthree. Both datasets provide explicit, verifiable intermediate states, allowing us to separately evaluate visual-state prediction and downstream reasoning.

Our experiments show that \modelname outperforms the fine-tuned backbone by 16.30--39.25 percentage points across \lanterntwo and \lanternthree. Replacing or corrupting its generated visual states substantially reduces performance, indicating that the model actively uses these states for reasoning. Training analyses further reveal that generated visual states become useful only after reaching sufficient quality, and that joint training from scratch fails under the tested setting. Moreover, a lightweight four-layer generation branch performs comparably to its full-depth counterpart. Together, these results suggest that effective visual reasoning requires not only generating informative visual states but also learning to use them. More broadly, our controlled study provides evidence that internal world modeling can enable VLMs to reason across visual and textual spaces, a capability that supports spatial reasoning and could ultimately serve as a foundation for embodied agents.

\section{Related Work} \label{sec:related}

\paragraph{Visual Reasoning with VLMs}
Vision-language models are first trained to reason in textual space, where model generates long chain-of-thought textual reasoning to solve a problem \citep{huang2026visionr1,r1-onevision2025,chart-r1,zha2026vision,DBLP:journals/corr/abs-2508-09804}. Previous work usually evaluate their model on STEM, charts or common sense related benchmarks, e.g., MathVista \citep{lu2024mathvista}, MMMU \citep{yue2024mmmu}, ChartQA \citep{masry2022chartqa}. However, not all vision-related tasks are well suited to textual reasoning. Prior work suggests that visuospatial tasks involving transformation, localization, or state tracking benefit more from visual intermediates or pixel-space operations than from purely textual reasoning \citep{larkin1987diagram,hu2024visual,su2026pixel}.

\paragraph{Interleaved Visual-Textual Reasoning} Prior work represents intermediate visual states in several forms and generates them through different mechanisms. Visual Sketchpad \citep{hu2024visual} and Pixel Reasoner \citep{su2026pixel} use external tools to edit input images (e.g., by zooming or cropping), treating the resulting images as intermediate states. MindJourney \citep{yang2026mindjourney} and DreamPlan \citep{jia2026dreamplan} invoke external world models to simulate future scenarios. Other methods use unified models, such as Anole \citep{chern2024anole} and BAGEL \citep{bagel2025emerging}, to generate intermediate reasoning images \citep{chern2025twgi,gu2026thinkmorph}. Mirage \citep{mirage2026}, LVR \citep{li2026lvr}, and Monet \citep{wang2026monet} train base VLMs to produce continuous visual latents, whereas LatentUM \citep{jin2026latentum} produces discrete visual tokens. Recent analyses, however, question whether models with interleaved visual-textual reasoning causally depend on these visual intermediates. \citet{viveiros2026s} find that removing or corrupting latent visual tokens often has little effect, attributing this behavior to redundant intermediate supervision that enables latent bypass, inference-time representation collapse, and a substantial oracle-generation gap.

\begin{figure}[tbp]
    \centering
    \includegraphics[width=\textwidth]{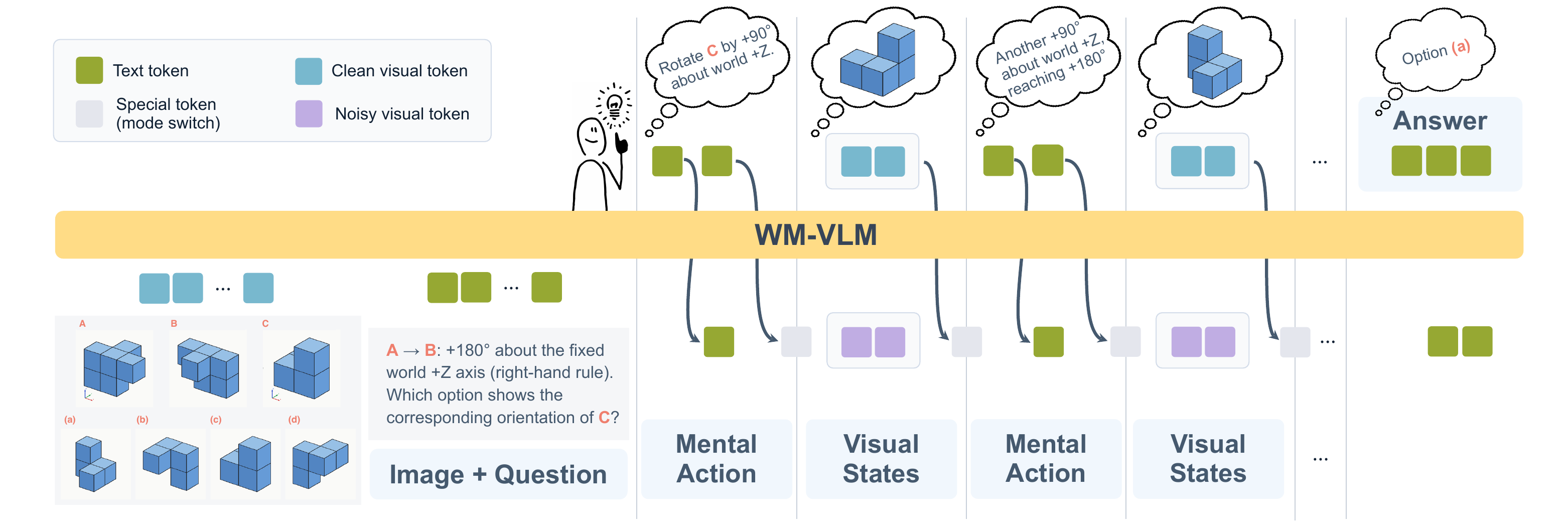}
    \caption{\modelname performs interleaved visual-textual reasoning by alternating between textual mental actions and generated visual states, before producing the final answer.}
    \label{fig:model-overview}
\end{figure}

\section{Method}
\subsection{Task Generation}
To study internal world models in interleaved visual-textual reasoning, we require datasets to meet three criteria. First, solving each task should require predicting a new visual state. Second, each example should include a question, an input image, an interleaved sequence of textual actions and visual states, and a final answer. Third, both the intermediate states and the final answer should be verifiable. However, datasets with all three properties remain scarce.

Following \citet{viveiros2026s}, we programmatically construct 2D and 3D spatial-reasoning datasets that satisfy these criteria. Each example contains an interleaved visual-textual reasoning trace in which intermediate visual states are needed to answer a verifiable multiple-choice question.

Each sample $s^{(i)}$ in the dataset $\mathcal{D}$ is represented as
\begin{equation}
s^{(i)} = \left(Q^{(i)}, I^{(i)}_q, R^{(i)}, A^{(i)}\right),
\end{equation}
where $Q^{(i)}$ is the initial question and $I^{(i)}_q$ is the corresponding question image. The reasoning trace is denoted as $R^{(i)} = \left((T_1, I_1)^{(i)}, (T_2, I_2)^{(i)}, ..., (T_n, I_n)^{(i)}, T^{(i)}_{n+1}\right)$, where $T_1$ is the first mental action on the original state of the image $I^{(i)}_q$, and $I_{1}$ is the resulted visual state after applying the mental action $T_1$. $I_{j}$ is the resulted visual state after applying the mental action $T_j$ on the previous visual state $I_{j-1}$, where $j > 1$. The textual summary of the interleaved visual and textual reasoning is denoted as $T^{(i)}_{n+1}$. The final answer is denoted as $A^{(i)}$.

\subsection{Interleaved Visual-Textual Reasoning with World Model}
We formulate interleaved visual-textual reasoning as a Markov process (Figure \ref{fig:model-overview}). Let $E$ denote the vision encoder, with $z_q=E(I_q)$ and $z_j=E(I_j)$ denoting the continuous visual embeddings of the initial image and the intermediate reasoning images, respectively. Before step $j$, the reasoning state $H_j$ contains the question and
the complete interleaved history, with $H_1=(Q,z_q)$:
\begin{equation}
    H_j=\left(Q,z_q,(T_1,z_1),\ldots,(T_{j-1},z_{j-1})\right).
\end{equation}
At each step, the policy model $\pi_\theta$ generates a piece of text including the mental action $T_j$. An internal world model $p^{\text{wm}}_\theta$ then predicts the resulting visual state in the embedding space:
\begin{equation}
    T_j\sim\pi_\theta(\cdot\mid H_j),
    \quad
    \hat{z_j}\sim p^{\text{wm}}_\theta(\cdot\mid H_j,T_j).
\end{equation}
During training, $E(I_j)$ provides the target embedding; during inference, the internal world model generates the embedding directly. The reasoning state is then updated as
\begin{equation}
    H_{j+1}=H_j\oplus(T_j,\hat{z_j}),
\end{equation}
where $\oplus$ denotes sequence concatenation. Thus, subsequent textual reasonings (including mental actions) are conditioned on the predicted visual outcomes of earlier steps. The model performs reasoning in this way after $n$ steps. Then the model generates the textual summary $T_{n+1}$ and the final answer $A$ from the accumulated reasoning state.

\begin{figure}[htbp]
    \centering
    \includegraphics[width=0.36\linewidth]{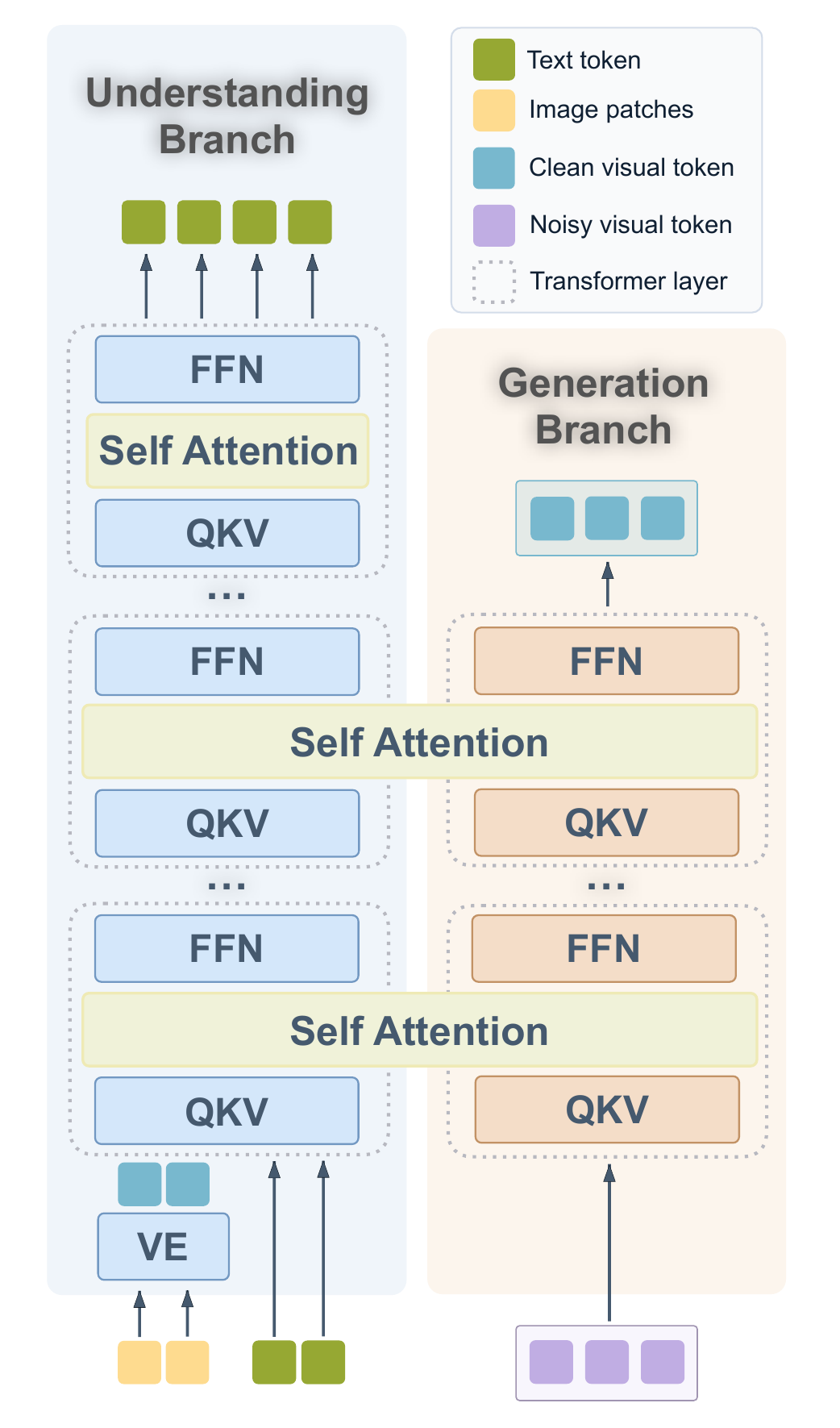}
    \caption{Architecture of \modelname, instantiated with our light Mixture-of-Transformers (light MoT) design. The understanding branch, comprising the vision encoder and language decoder, is initialized from a pretrained VLM. The shallow generation branch contains \(k\) layers, each paired with a corresponding layer in a consecutive \(k\)-layer block of the understanding branch. Text and clean visual tokens are routed to the understanding branch, whereas noisy visual tokens are routed to the generation branch. All tokens interact through global self-attention.}
    \label{fig:model-architecture}
\end{figure}

\subsection{VLM with An Internal World Model} \label{sec:vlm-w-an-internal-wm}
We use the Mixture-of-Transformer (MoT) architecture \citep{liang2024mixture} that accomodates both visual understanding and world modeling capabilities. Our model contains an understanding branch and a generation branch. The understanding branch is initialized with a pre-trained vision-language model (e.g., Qwen2.5-VL) and the generation branch weights are intialized from the corresponding understanding branch layer.

Each token $x_i$ is routed according to its modality $m_i$, where $m_i\in\{\text{text}, \text{clean image}, \text{noisy image}\}$. Text and clean image tokens are routed to the understanding branch, whereas noisy image tokens are routed to the generation branch. Clean image tokens comprise the encoded question-image tokens and the visual tokens generated by the generation branch. The generation branch transforms noisy image tokens into clean image tokens.

In practice, text and clean image tokens share one set of parameters, whereas noisy image tokens are processed by a separate parameter set in the generation branch. After generation, each noisy image token is replaced by its corresponding clean image token. Attention is computed globally across tokens from both branches.

To improve training efficiency, we introduce a \textbf{light MoT} architecture (Figure \ref{fig:model-architecture}) that retains the full understanding branch but uses a shallow generation branch. We align its $k$ generation layers with $k$ consecutive layers of the understanding branch. Generation tokens are processed only in these aligned layers, where the model applies global attention across tokens from both branches. We set $k=4$ by default.

\subsection{Two-stage Training}
Training proceeds in two stages. In the first stage, the generation branch learns to predict the next visual state conditioned on a mental action. In the second stage, the VLM learns to use the generated visual latents for downstream reasoning.

In Stage 1, we freeze the entire understanding branch, including the vision encoder and language decoder, and train only the generation branch. In Stage 2, we keep the vision encoder frozen and jointly train the language decoder and generation branch.

We optimize the generation branch with a rectified flow objective \citep{rectifiedflow2023}
and the understanding branch with a cross-entropy objective:
\begin{equation}
    \mathcal{L}
    = \alpha \mathcal{L}_{\mathrm{flow}}
    + \beta \mathcal{L}_{\mathrm{CE}},
\end{equation}
where $\alpha$ and $\beta$ balance the two scalar loss terms.

Specifically, given a clean target latent $z^{(1)}$ and conditioning
information $c$, we construct an interpolated latent
$z^{(t)} = (1-t)\epsilon + t z^{(1)}$, where
$\epsilon \sim \mathcal{N}(0,I)$ and $t \in [0,1]$.
Thus, $t=0$ corresponds to noise and $t=1$ to the clean target latent.
For visual latents, parenthesized superscripts denote flow time, while subscripts denote reasoning-step indices. The generation branch predicts the velocity along this path, yielding the rectified flow objective:
\begin{equation}
    \mathcal{L}_{\mathrm{flow}}
    =
    \mathbb{E}_{z^{(1)},c,t,\epsilon}
    \left[
        \left\|
            v_{\theta}^{\mathrm{gen}}(z^{(t)},t;c)
            - (z^{(1)}-\epsilon)
        \right\|_2^2
    \right],
\end{equation}
where $v_{\theta}^{\mathrm{gen}}$ denotes the velocity field
predicted by the generation branch. Here, $z^{(1)}$ is the visual embedding obtained by passing the intermediate reasoning image through the vision encoder, and the conditioning information $c$ includes the question and previous reasoning steps. In our experiments, we set $\alpha=1, \beta=0$
in Stage~1, and $\alpha=0, \beta=1$ in Stage~2.

At inference time, we generate $\hat z_j$ by integrating the learned velocity field conditioned on $c_j=(H_j,T_j)$. We then append $\hat z_j$ and $T_j$ to the reasoning history to condition subsequent steps. We provide the numerical integration procedure in Appendix~\ref{sec:flow-sampling}.

\section{Experiments}
\subsection{Implementation Details}
\subsubsection{Data Construction}
We use programs to generate the interleaved visual-textual reasoning dataset. It guarantees that each sample has at least one intermediate reasoning image that is essential for answering the final question. Specifically, following \citet{viveiros2026s}, we intialize 2D and 3D shapes with random number of atomic squares and cubes, respectively. We then apply a series of rotation to create different views of these shapes. Each question first presents two views of the same reference shape that illustrate a rotation. The model is then shown a new shape and asked to identify the view produced by applying the same rotation. A model capable of mental rotation should solve the task with only a short reasoning chain. We name the 2D rotation dataset as \textit{\lanterntwo} and the 3D rotation dataset as \textit{\lanternthree}. \lanterntwo contains 4k training samples, 400 in-distribution (ID) and 500 out-of-distribution (OOD) test cases, respectively. \lanternthree contains 16k training samples. Because 3D rotation is more complex and harder than 2D rotation. We include 400 easy in-domain samples (\lanternthree-SC-ID), where the same cube shape and count appear in the training data. Additional 400 hard in-domain samples (\lanternthree-C-ID) include seen cube count but unseen cube shapes. The 500 OOD test cases include cube shapes or counts that never appear in the training data. More details of building \lanterntwo and \lanternthree are shown in Appendix \ref{sec:build-lantern}. We also train on ThinkMorph-SpatialNavigation \citep{gu2026thinkmorph}, which combines state prediction with planning. Given a map, a starting point, and a destination, the model must find a path that avoids ice holes. At each step, it generates a visual state representing the agent's current position and plans the next move. Following \citet{gu2026thinkmorph}, we use the maze navigation task in the VSP benchmark \citep{wu2024vsp} as our testbed.

\begin{table}[htbp]
    \centering
    \caption{Performance comparison of methods with different reasoning types and training objectives. All models in this table are fine-tuned on \lanterntwo or \lanternthree, except Qwen2.5-VL-7B-Inst. Results are accuracy (\%); $\Delta$ denotes the absolute improvement of \modelname over Qwen2.5-VL-7B-Instruct SFT in percentage points. AR, FM, CE, and MSE denote autoregressive, flow matching, cross-entropy, and mean squared error, respectively. *Fine-tuned BAGEL (ThinkMorph) fails to emit images and generate answers on \lanternthree, resulting in 0\% accuracy.}
    \label{tab:world-modeling-objective}
    \setlength{\tabcolsep}{5pt}
    \resizebox{0.95\linewidth}{!}{%
    \begin{tabular}{@{}lllrrrrr@{}}
        \toprule
        \multirow[c]{2}{*}{Method} & \multirow[c]{2}{*}{Visual Reasoning Type} & \multirow[c]{2}{*}{Objective} & \multicolumn{2}{c}{\lanterntwo} & \multicolumn{3}{c}{\lanternthree} \\
        \cmidrule(lr){4-5} \cmidrule(lr){6-8}
        & & & \multicolumn{1}{c}{ID} & \multicolumn{1}{c}{OOD} & \multicolumn{1}{c}{SC-ID} & \multicolumn{1}{c}{C-ID} & \multicolumn{1}{c}{OOD} \\
        \midrule
        Qwen2.5-VL-7B-Inst. & (Text only) & AR-CE & 22.00 & 23.80 & 21.25 & 23.00 & 21.40 \\
        + SFT & (Text only) & AR-CE & 48.25 & 46.20 & 71.75 & 44.50 & 55.50 \\ \midrule
        LatentUM & Discrete Visual Tokens & AR-CE & 29.25 & 23.00 & 58.50 & 37.75 & 39.00 \\
        Mirage & Continuous Visual Tokens & AR-Cosine & 47.75 & 43.00 & 66.25 & 46.50 & 49.80 \\
        ThinkMorph & Generated Image & FM-MSE & 22.50 & 21.60 & 0.00* & 0.00* & 0.00* \\
        \midrule
        \textbf{\modelname (Ours)} & Continuous Visual Tokens & FM-MSE & \textbf{87.50} & \textbf{75.00} & \textbf{91.00} & \textbf{66.25} & \textbf{71.80} \\
        $\Delta_{+\text{SFT}\rightarrow\text{Ours}}$ & / & / & +39.25 & +28.80 & +19.25 & +21.75 & +16.30 \\
        \bottomrule
    \end{tabular}%
    }
\end{table}

\subsubsection{Model and Training Details}
We use the light MoT architecture introduced in Section \ref{sec:vlm-w-an-internal-wm}. The understanding branch in light MoT is intialized with Qwen2.5-VL-7B-Instruct \citep{qwen2.5-vl}.
For each dataset, we start with freezing the understanding branch and only train the generation branch until convergence. Then we unfreeze the understanding branch and jointly train both branches. We use an online training setting in which the understanding branch receives visual tokens generated by the generation branch rather than ground truth tokens. We keep the vision encoder frozen during the entire training. Hyperparameters are shown in Table \ref{tab:training-hyperparameters}, in the appendix.

For baselines, we first compare \modelname with a supervised fine-tuned Qwen2.5-VL-7B-Instruct model (Qwen2.5-VL-7B-Instruct SFT), as it provides the most direct comparison. Qwen2.5-VL is a VLM trained with a standard autoregressive (AR) objective and optimized using cross-entropy (CE) loss. For fairness, we optimize Qwen2.5-VL with the same steps as \modelname. Next, we include LatentUM \citep{jin2026latentum}, which generates discrete visual tokens autoregressively. We use their pre-trained model (LatentUM-Base\footnote{\url{https://huggingface.co/SJTU-DENG-Lab/LatentUM-Base}}) and conduct continual training for learning the world model and training the reasoning capability. We also include Mirage \citep{mirage2026}, which generates continuous visual tokens autoregressively. We reuse their two-stage training approach on our datasets. Finally, we include ThinkMorph \citep{gu2026thinkmorph}, which generates image pixels and is trained with a flow-matching loss. ThinkMorph is built on the unified model BAGEL, which has already been trained on large-scale interleaved visual-textual data. Following \citet{gu2026thinkmorph}, we simply finetune BAGEL on each of the datasets and then do evaluation. All experiments use 8 H200 GPUs if not otherwise specified.

\subsection{Results and Analysis}
Vision-language models learn from image-text data using a next-token prediction objective, which does not take full advantage of the rich supervision provided by visual inputs. Consequently, VLMs fail to learn effective internal world models that can predict future visual states conditioned on the current state and mental action.

\begin{table}[htbp]
    \centering
    \setlength{\belowcaptionskip}{2pt}
    \caption{Relationship between visual-token retrieval quality and answer accuracy. Cosine denotes the average cosine similarity between the generated and ground truth visual embeddings. Retrieval top-1 denotes the percentage of times the correct ground truth embedding is retrieved. Accuracy-\ding{51} is the accuracy when the retrieval is correct, while Accuracy-\ding{55} is the accuracy when the retrieval is wrong. $\Delta$ denotes the absolute change relative to standard accuracy. The $\phi$ coefficient is computed between two binary indicators: whether the correct ground-truth embedding is retrieved and whether the final answer is correct.}
    \label{tab:retrieval-quality-answer-accuracy}
    \setlength{\tabcolsep}{5pt}
    \resizebox{0.85\textwidth}{!}{%
    \begin{tabular}{@{}lrrrrr@{}}
        \toprule
        Metric & \lanterntwo-ID & \lanterntwo-OOD & \lanternthree-SC-ID & \lanternthree-C-ID & \lanternthree-OOD \\
        \midrule
        Cosine & 0.9347 & 0.7523 & 0.8511 & 0.7921 & 0.7746 \\
        Retrieval top-1 & 96.50 & 53.00 & 65.75 & 20.75 & 13.40 \\
        Standard accuracy & 87.50 & 75.00 & 91.00 & 66.25 & 71.80 \\
        \midrule
        Accuracy-\ding{51} & 89.38 & 79.25 & 99.24 & 92.77 & 86.57 \\
        $\Delta$ & +1.88 & +4.25 & +8.24 & +26.52 & +14.77 \\ \midrule
        \addlinespace[2pt]
        Accuracy-\ding{55} & 35.71 & 70.21 & 75.18 & 59.31 & 69.52 \\
        $\Delta$ & -51.79 & -4.79 & -15.82 & -6.94 & -2.28 \\ \midrule
        $\phi$ correlation & 0.298 & 0.104 & 0.399 & 0.287 & 0.129 \\
        \bottomrule
    \end{tabular}%
    }
\end{table}

We run a comparative study to show that adding a world modeling objective is essential. Starting with \lanterntwo, we first train the generation branch in \modelname with the flow matching loss, enabling its world modeling capability (Stage 1). Then we train both the understanding and generation branch to unlock the interleaved visual-textual reasoning capability (Stage 2). We evaluate the final checkpoint on both \lanterntwo-ID and \lanterntwo-OOD.

\paragraph{World modeling objective boosts performance on task requires imagination.} Training dynamics and evaluation results are shown in Figure \ref{fig:tetris-id-ood-accuracy-cosine} and Table \ref{tab:world-modeling-objective}, respectively. Figure \ref{fig:tetris-id-ood-accuracy-cosine} shows a clear performance transition at around 25k training steps. Meanwhile, the cosine similarity between the generated visual states and the ground-truth embedding also increases more rapidly around the same steps. Our assumption is that the model is only able to benefit from the world model once its generation capability exceeds a certain threshold. Qwen2.5-VL-7B-Instruct SFT demonstrates similar training dynamics before step 25k and exhibits downstream performance comparable to \modelname. However, its performance plateaus after step 25k, while \modelname enters the transition and ultimately outperforms Qwen2.5-VL-7B-Instruct SFT by 39.25 points on \lanterntwo-ID and 28.8 points on \lanterntwo-OOD.

Similar performance gains are observed on the harder \lanternthree evaluation sets, where \modelname outperforms Qwen2.5-VL-7B-Instruct SFT by 19.25 points on \lanternthree-SC-ID, 21.75 points on \lanternthree-C-ID, and 16.3 points on \lanternthree-OOD. These results show that adding a world modeling objective is essential for solving spatial reasoning tasks that require imagination. \modelname also outperforms other visual latent reasoning and interleaved visual-textual reasoning methods on \lanterntwo and \lanternthree.

\begin{table}[htbp]
    \centering
    \begin{minipage}[t]{0.44\linewidth}
        \vspace{0pt}
        \centering
        \caption{Ablations on alternating generated image pixels on \lanterntwo. We report accuracy (\%). $\Delta$ denotes the change relative to standard inference with generated pixels.}
        \label{tab:pixel-reconstruction-evaluation}
        \setlength{\tabcolsep}{5pt}
        \resizebox{\linewidth}{!}{%
        \begin{tabular}{@{}lccc@{}}
            \toprule
            Metric
                & \shortstack{all-white \\ pixel}
                & \shortstack{shuffle \\ $19{\times}19$ patches}
                & \shortstack{Default} \\
            \midrule
            \lanterntwo-ID & 49.25 & 48.25 & 47.25 \\
            $\Delta$ & +2.00 & +1.00 & -- \\
            \midrule
            \lanterntwo-OOD & 46.60 & 46.20 & 46.20 \\
            $\Delta$ & +0.40 & 0.00 & -- \\
            \bottomrule
        \end{tabular}%
        }
    \end{minipage}%
    \hfill
    \begin{minipage}[t]{0.54\linewidth}
        \vspace{0pt}
        \centering
        \caption{Ablations on alternating generated visual tokens on \lanterntwo. Results are accuracy (\%); $\Delta$ denotes the change relative to evaluation with the default setup.}
        \label{tab:visual-token-manipulation-test}
        \setlength{\tabcolsep}{5pt}
        \resizebox{\linewidth}{!}{%
        \begin{tabular}{@{}lcccc@{}}
            \toprule
            Metric
                & \shortstack{all-zero \\ embedding}
                & \shortstack{randomized \\ tokens}
                & \shortstack{shuffle \\ visual tokens}
                & \shortstack{Default} \\
            \midrule
            \lanterntwo-ID & 14.00 & 23.50 & 81.50 & 87.50 \\
            $\Delta$ & -73.50 & -64.00 & -6.00 & -- \\
            \midrule
            \lanterntwo-OOD & 15.40 & 20.80 & 72.60 & 75.00 \\
            $\Delta$ & -59.60 & -54.20 & -2.40 & -- \\
            \bottomrule
        \end{tabular}%
        }
    \end{minipage}
\end{table}

\begin{table}[htbp]
    \centering
    \caption{Ablations on alternating generated visual tokens on \lanternthree. Results are accuracy (\%); $\Delta$ denotes the change relative to evaluation with the default setup.}
    \label{tab:tetris-3d-visual-token-evaluation}
    \setlength{\tabcolsep}{5pt}
    \resizebox{0.6\linewidth}{!}{%
    \begin{tabular}{@{}lcccc@{}}
        \toprule
        Metric
            & \shortstack{all-zero \\ embedding}
            & \shortstack{randomized \\ tokens}
            & \shortstack{shuffle \\ visual tokens}
            & \shortstack{default} \\
        \midrule
        \lanternthree-SC-ID & 15.50 & 23.25 & 81.00 & 91.00 \\
        $\Delta$ & -75.50 & -67.75 & -10.00 & -- \\
        \midrule
        \lanternthree-C-ID & 16.00 & 25.00 & 60.25 & 66.25 \\
        $\Delta$ & -50.25 & -41.25 & -6.00 & -- \\
        \midrule
        \lanternthree-OOD & 14.40 & 20.40 & 64.20 & 71.80 \\
        $\Delta$ & -57.40 & -51.40 & -7.60 & -- \\
        \bottomrule
    \end{tabular}%
    }
\end{table}

\paragraph{Do generated visual tokens improve reasoning?} The model may learn shortcuts that allow it to answer questions without relying on the generated visual tokens. To test this possibility, we conduct a retrieval experiment to evaluate the contribution of these tokens. As we have all the intermediate reasoning images in the eval set, we first encode these images to obtain the ground truth visual embeddings. Then, we get the \modelname's generated visual embeddings, and compute the cosine similarity between the generated and ground truth embeddings. We select the top-1 retrieved ground truth embedding for each generated embedding. As shown in Table \ref{tab:retrieval-quality-answer-accuracy}, higher retrieval rate leads to higher answer accuracy. Correlation analysis shows that the retrieval quality is positively correlated with the answer accuracy.

\begin{figure}[htbp]
    \centering
    \includegraphics[width=0.92\textwidth]{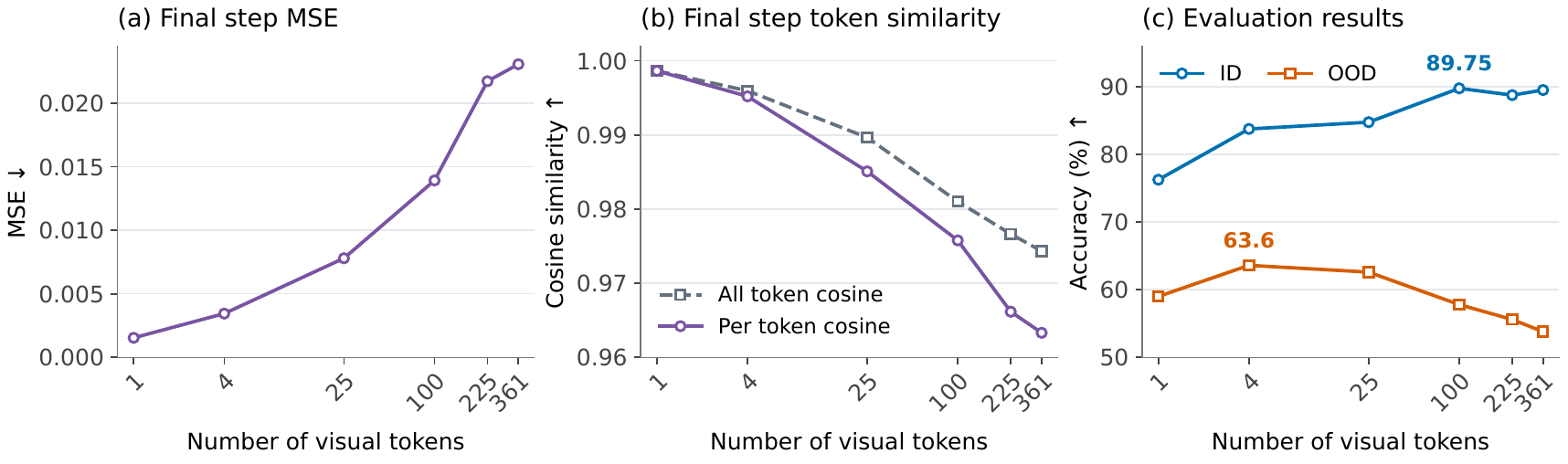}
    \caption{Effect of visual-token count on model performance. We report final-step MSE, training-time token similarity, and performance on \lanterntwo-ID and \lanterntwo-OOD. MSE compares the generated and ground-truth visual embeddings. All-token cosine is computed after concatenating all tokens in each block, whereas per-token cosine averages the similarities between corresponding generated and ground-truth tokens.}
    \label{fig:latent-count-ablation}
\end{figure}

Further, we manipulate the generated visual states to see how it affects the final performance. We conduct three types of manipulation: (1) replace the generated visual tokens with all-zero embeddings, (2) replace the generated visual tokens with randomized embeddings, and (3) shuffle the generated visual tokens. As shown in Table \ref{tab:visual-token-manipulation-test} and \ref{tab:tetris-3d-visual-token-evaluation}, replacing the visual tokens with all-zero embeddings or randomized  embeddings leads to a significant drop in accuracy. This indicates that the generated visual tokens are non-trivial and essential for assisting the model to answer the final question. It is expected that shuffling the visual tokens has a smaller impact, because visual tokens do bi-directional attention with each other in the generation branch. They already have positional information and shuffling tokens does not change that.

\begin{figure}[htbp]
    \centering
    \includegraphics[width=0.92\textwidth]{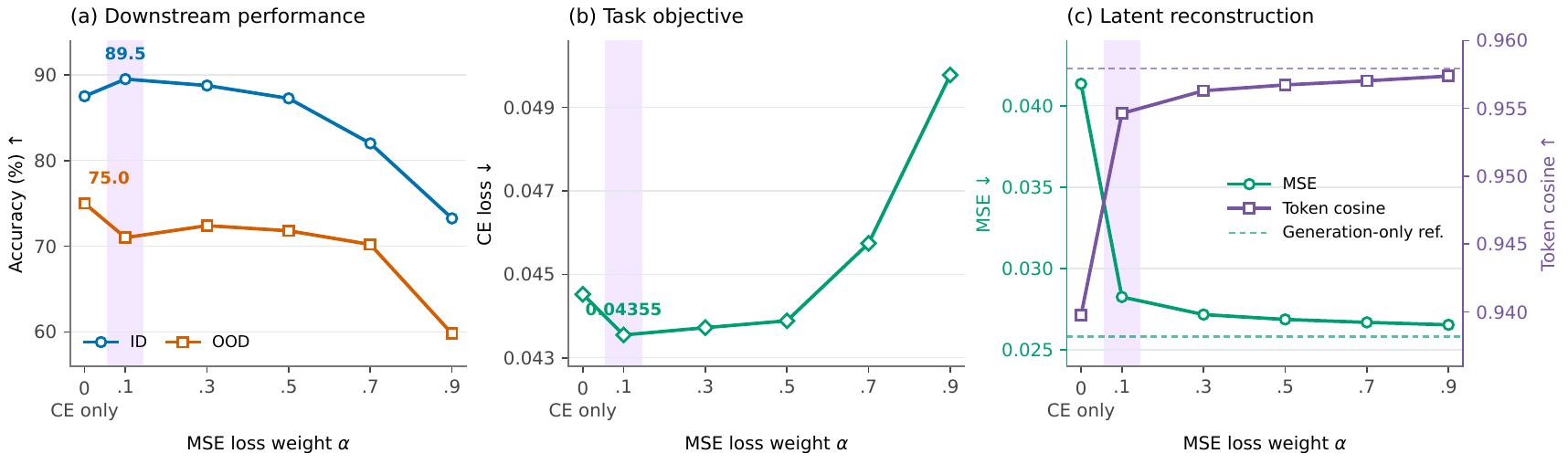}
    \caption{Effect of the ratio between cross-entropy and flow-matching losses on model performance.}
    \label{fig:loss-weight-ablation}
\end{figure}

\paragraph{Does the number of visual tokens affect model performance?} As the size of the reasoning images in the dataset is 512$\times$512, the default number of visual tokens is 361 (19$\times$19) in our experiments. We conduct an ablation study to see how the number of visual tokens affects model performance. We change the number of visual tokens by simply resizing the reasoning images. As shown in Figure \ref{fig:latent-count-ablation}, fewer visual tokens leads to lower MSE and higher cosine similarity. However, the final answer accuracy does not show the same trend. ID performance increases as the number of visual tokens increases, while OOD performance peaks at 4 ($2\times2$) visual tokens.

\begin{figure}[htbp]
    \centering
    \includegraphics[width=0.7\textwidth]{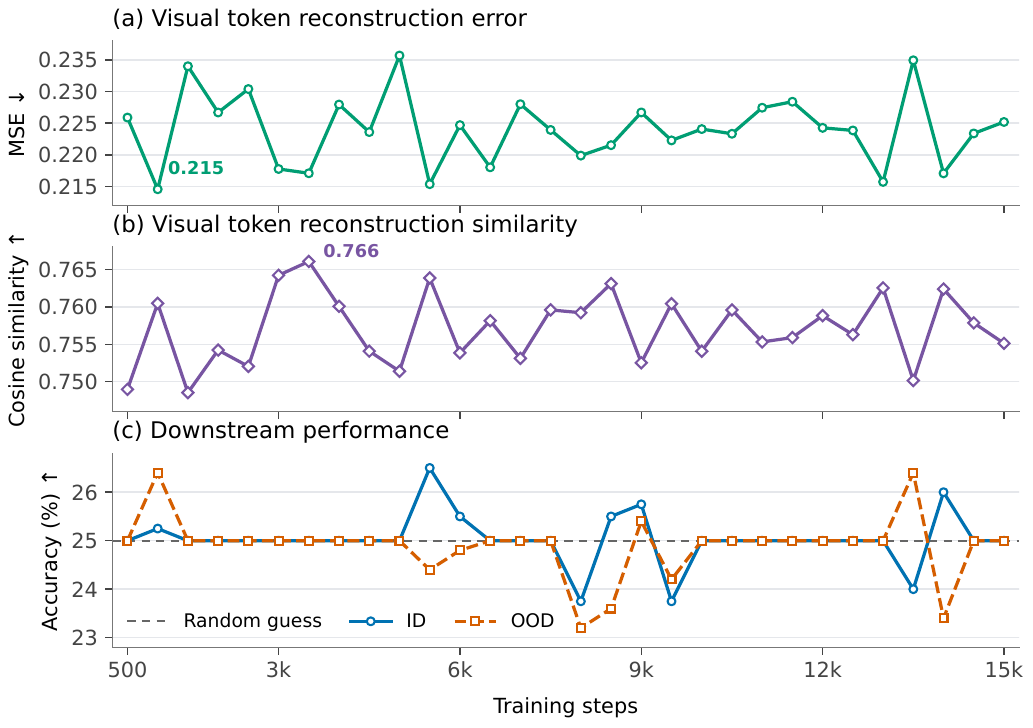}
    \caption{Training curves for joint training with loss ratio $\alpha=0.1$ and $\beta=0.9$.}
    \label{fig:single-stage-training-curves}
\end{figure}

\paragraph{What is the optimal loss ratio in Stage 2?} By default, we only use the cross-entropy loss in Stage 2 ($\alpha=0, \beta=1$), to optimize both the understanding and generation branches. To see how the ratio between the cross-entropy and flow-matching losses affects model performance, we conduct an ablation study by varying $\alpha$ and $\beta$. Starting from the same checkpoint in Stage 1, we train the model with different loss ratios in Stage 2. We set $\alpha=\{0.1, 0.3, 0.5, 0.7, 0.9\}$, with $\beta=1-\alpha$. As shown in Figure \ref{fig:loss-weight-ablation}, larger weight of the flow-matching loss in Stage 2 improves the generated visual token quality. But the improved visual token quality does not necessarily lead to better downstream performance. We observe that the best overall performance is achieved when $\alpha=0.1$ and $\beta=0.9$, showing cross-entropy dominating the final downstream reasoning capability.

\begin{figure}[htbp]
    \centering
    \includegraphics[width=0.8\textwidth]{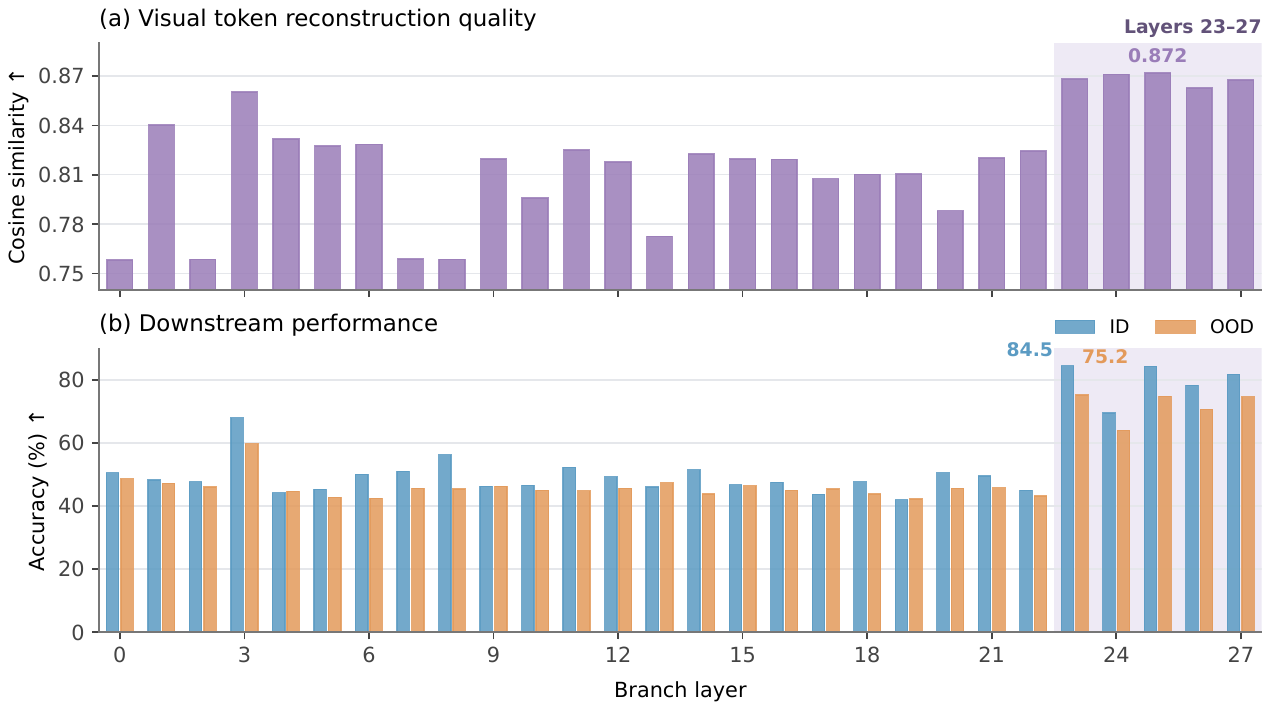}
    \caption{Layer-wise analysis of visual latent quality and downstream performance in the VLM.}
    \label{fig:ablation-one-layer-generation-branch}
\end{figure}

\paragraph{Is two-stage training necessary?}
Our two-stage training strategy first trains the model to predict visual tokens and then teaches it to reason using these tokens. A natural alternative is to train both branches jointly from scratch using a weighted combination of the two training objectives. To evaluate this alternative, we perform single-stage training with $\alpha=0.1$ and $\beta=0.9$, the loss weights that yield the best performance in our preceding experiments. As shown in Figure~\ref{fig:single-stage-training-curves}, the accuracy of the single-stage model remains near random chance throughout training. In contrast, Qwen2.5-VL-7B-Instruct SFT and \modelname both achieve over 40\% accuracy within a comparable number of training steps. These results suggest that, under our training setup, jointly learning visual-token prediction and visual-token-based reasoning from scratch is ineffective. The staged curriculum is critical because the model must first learn a representation of the visual world before it can reason effectively with the generated visual tokens.

\begin{figure}[htbp]
    \centering
    \includegraphics[width=0.9\textwidth]{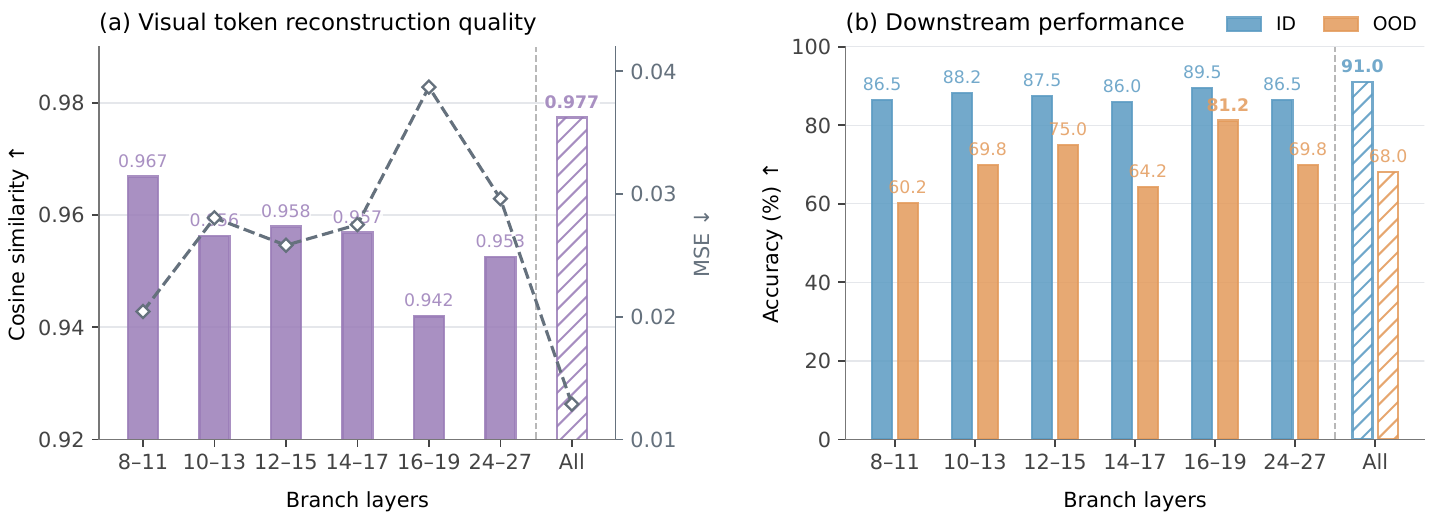}
    \caption{Effect of the layer-window selection on model performance. For example, "8-11" means the generation branch is connected to the understanding branch from layer 8 to layer 11.}
    \label{fig:layer-window-ablation}
\end{figure}

\paragraph{How many generation layers are needed, and where should they be placed?} The light MoT architecture (Figure \ref{fig:model-architecture}) allows the generation branch to contain any number of layers and to align with any consecutive block of layers in the understanding branch. We study if alternating the number of generation layers and their placement affects model performance. First, we intialize \modelname with 1-layer generation branch and train it on \lanterntwo. The results are shown in Figure \ref{fig:ablation-one-layer-generation-branch}, where aligning generation layer with the top-most (23-27)layers in the understanding branch, the model achieves the best performance. Aligning with other layers shows lower performance, except layer-3, where there is a  performance spike. Next, we scan the understanding branch with a 4-layer generation branch. As shown in Figure \ref{fig:layer-window-ablation}, when the number of generation layers increases, the performance gap between different placements becomes smaller. We attribute this to the increased model capacity that can mitigate the informational loss from different understanding branches. On the other hand, using the same number of generation layers with understanding layers (the original MoT setting) does not show significant performance gain.

\paragraph{Can we use pixels as the intermediate visual states?} By default, we use the generated visual tokens (embeddings), which are aligned with the vision encoder output, as the intermediate visual states. But it is also possible to use the generated pixels to represent the intermediate reasoning state. We conduct an ablation study by simply adding an MLP on top of the generated visual tokens to reconstruct the pixels. We train the pixel-generation model with the pixel MSE loss, with the same optimizing steps as \modelname. Experiments result in Table \ref{tab:pixel-reconstruction-evaluation} and \ref{tab:visual-token-manipulation-test} show that given the same compute budget, reasoning with the generated visual tokens yields better performance than the pixels.

\section{Conclusion}
To enable VLMs to reason about space through visual imagination, we introduce \modelname, a VLM equipped with an internal world model. We construct datasets with verifiable intermediate visual states and use them to study how world modeling affects interleaved visual-textual reasoning. Our experiments show that \modelname improves spatial reasoning performance and relies on its generated visual tokens to produce final answers. Ablations identify the number of visual tokens, the balance between cross-entropy and flow-matching losses, and two-stage training as important design choices. More broadly, internal world models may enable VLMs to reason jointly in language and visual space.

\bibliography{references}
\bibliographystyle{assets/plainnat}

\clearpage
\beginappendix
\section{Implementation Details}
\subsection{VLM with an internal world model}
In \modelname, each token's query, key, and value are computed as follows:
\begin{equation}
{
Q_i=x_iW_Q^{m_i},\quad
K_i=x_iW_K^{m_i},\quad
V_i=x_iW_V^{m_i},
}
\end{equation}
where $W_Q^{m_i}$, $W_K^{m_i}$, and $W_V^{m_i}$ are the learnable weight matrices for the query, key, and value projections of modality $m_i$, respectively.

The attention output $a_i$ for token $x_i$ in a transformer layer is:
\begin{equation}
a_i=
\left[
\operatorname{softmax}\!\left(
\frac{QK^\top}{\sqrt{d_k}}
\right)V
\right]_i
W_O^{m_i},
\end{equation}
where $W_O^{m_i}$ is the learnable weight matrix for the output projection of modality $m_i$.
\subsection{Rectified Flow Sampling} \label{sec:flow-sampling}
At inference time, we generate the estimated visual state $\hat z_j$ conditioned on $c_j=(H_j,T_j)$ by integrating the predicted velocity field from noise ($t=0$) to the clean latent ($t=1$). Specifically, we initialize $\tilde z_j^{(0)}\sim\mathcal{N}(0,I)$ and use $K$ Euler steps with $t_k=k/K$:
\begin{equation}
    \tilde z_j^{(t_{k+1})}
    = \tilde z_j^{(t_k)}
    + \frac{1}{K}\,
    v_{\theta}^{\mathrm{gen}}\!\left(\tilde z_j^{(t_k)},t_k;c_j\right),
    \qquad k=0,\ldots,K-1.
\end{equation}
The final estimate is $\hat z_j=\tilde z_j^{(1)}$, which is appended to the reasoning history together with $T_j$ to condition subsequent reasoning steps. Here, $K$ denotes the number of Euler steps, while $t$ denotes the continuous flow time.
\subsection{Qwen2.5-VL SFT}
We use the supervised fine-tuned Qwen2.5-VL-7B-Instruct model as our baseline. Because \lanterntwo and \lanternthree contain interleaved visual-textual samples, it is not possible for Qwen2.5-VL to learn from the intermediate image (as it only gets supervision signal from text tokens). To make the comparison fair, we replace the intermediate image with a text token ''[IMAGINATION]'', and do ordinary SFT on the modified dataset. Alternatively, we keep the intermediate reasoning image and let it appear in the training. The experiment results are shown in Table~\ref{tab:qwen-sft-input-settings}, which suggests that SFT with the token ''[IMAGINATION]'' yields better performance. We use the former setting as the baseline for comparison with \modelname.

\begin{table}[htbp]
    \centering
    \caption{Accuracy (\%) of Qwen2.5-VL-7B-Instruct on \lanterntwo after SFT with intermediate reasoning images replaced by an \texttt{[IMAGINATION]} token or retained as raw interleaved images. Bold denotes the best accuracy across optimization steps for each setting and test split.}
    \label{tab:qwen-sft-input-settings}
    \small
    \setlength{\tabcolsep}{8pt}
    \renewcommand{\arraystretch}{1.08}
    \begin{tabular}{@{}ccccc@{}}
        \toprule
        \multirow{2}{*}{\shortstack{Optimization\\steps}}
            & \multicolumn{2}{c}{\texttt{[IMAGINATION]} token}
            & \multicolumn{2}{c}{Raw interleaved images} \\
        \cmidrule(lr){2-3} \cmidrule(l){4-5}
            & \multicolumn{1}{c}{ID} & \multicolumn{1}{c}{OOD}
            & \multicolumn{1}{c}{ID} & \multicolumn{1}{c}{OOD} \\
        \midrule
        2,500  & 25.50 & 25.40 & 28.25 & 22.80 \\
        5,000  & 47.75 & 45.20 & \textbf{30.00} & 24.20 \\
        7,500  & \textbf{49.50} & \textbf{45.60} & 24.75 & 23.00 \\
        10,000 & 44.50 & 44.40 & 25.50 & \textbf{25.00} \\
        \bottomrule
    \end{tabular}
\end{table}

\subsection{Training Hyperparameters}
We present the training hyperparameters in Table \ref{tab:training-hyperparameters}. All training tasks are conducted on 8 H200 GPUs.
\section{Data Construction}
\subsection{Building \lanterntwo and \lanternthree} \label{sec:build-lantern}
We use programs to generate the interleaved visual-textual reasoning dataset. Detailed algorithm is shown in Algorithm \ref{alg:lantern3d_simple}. The 2D rotation dataset is constructed in a similar way, except that we use squares instead of cubes.
\begin{algorithm}[ht]
\caption{Construction of the \lanternthree Dataset}
\label{alg:lantern3d_simple}
\begin{algorithmic}[1]
\Require Cube counts $\{4,5,6,7\}$, rotation axes $\{X,Y,Z\}$,
rotation angles $\{90^\circ,180^\circ,270^\circ\}$
\Ensure Training set $\mathcal{D}_{\mathrm{train}}$ and evaluation sets
$\mathcal{D}_{\mathrm{ID}},\mathcal{D}_{\mathrm{OOD}}$

\State $\mathcal{S}\gets\emptyset$
\For{$n\in\{4,5,6,7\}$}
    \State Enumerate all connected shapes consisting of $n$ cubes
    \State Remove shapes equivalent under proper 3D rotations
    \State Add the remaining canonical shapes to $\mathcal{S}$
\EndFor

\State Split $\mathcal{S}$ into seen shapes $\mathcal{S}_{\mathrm{seen}}$
and held-out shapes $\mathcal{S}_{\mathrm{OOD}}$
\Comment{Mirror-related shapes stay in the same split}

\State $\mathcal{Q}_{\mathrm{seen}}\gets\emptyset$,
$\mathcal{Q}_{\mathrm{OOD}}\gets\emptyset$

\For{each shape $S\in\mathcal{S}$}
    \For{each distinct initial orientation $P$ of $S$}
        \For{$a\in\{X,Y,Z\}$ and
             $\theta\in\{90^\circ,180^\circ,270^\circ\}$}
            \State Compute the rotated orientation
            $P'=R(a,\theta)P$
            \If{$P'\neq P$}
                \State Add $(S,P,a,\theta,P')$ to the corresponding
                seen or OOD configuration pool
            \EndIf
        \EndFor
    \EndFor
\EndFor

\State Split the seen configuration pool into disjoint
training and ID-evaluation pools

\For{each selected configuration $(S_C,P_C,a,\theta,P'_C)$}
    \State Select a reference shape $S_A$ with a different cube count
    \State Rotate $S_A$ by the same $(a,\theta)$ to produce Reference B
    \State Use $P'_C$ as the correct answer
    \State Sample three distinct rotated poses of $S_C$ as distractors
    \State Randomize the position of the four answer options
    \State Render Reference A, Reference B, Query C, and the four options
\EndFor

\State Construct $\mathcal{D}_{\mathrm{train}}$ from training configurations
\State Construct $\mathcal{D}_{\mathrm{ID}}$ from unseen configurations of seen shapes
\State Construct $\mathcal{D}_{\mathrm{OOD}}$ from held-out shapes

\end{algorithmic}
\end{algorithm}

\subsection{Data Samples}
\begin{figure}[htbp]
    \centering
    \begin{minipage}[t]{0.48\linewidth}
        \centering
        \textbf{(a) \lanterntwo}\\[0.25em]
        \includegraphics[width=\linewidth]{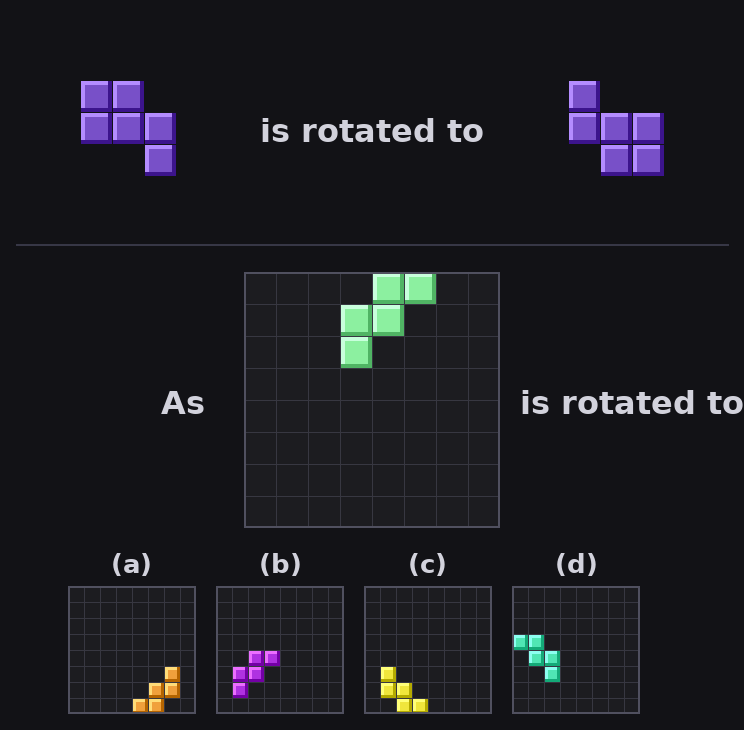}\\[0.25em]
        {\footnotesize
        \textbf{$T_1,T_2$:} Rotate the query shape $90^\circ$ clockwise twice.\\[-0.25em]
        $\downarrow$\\[-0.25em]
        }
        \includegraphics[width=0.34\linewidth]{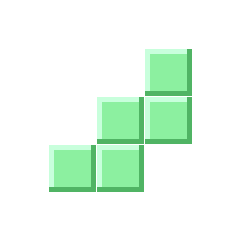}\\[-0.15em]
        {\footnotesize \textbf{Final visual state}\\[-0.1em]
        Compare with the options. \textbf{Answer: (a)}}
    \end{minipage}
    \hfill
    \begin{minipage}[t]{0.48\linewidth}
        \centering
        \textbf{(b) \lanternthree}\\[0.25em]
        \includegraphics[width=\linewidth]{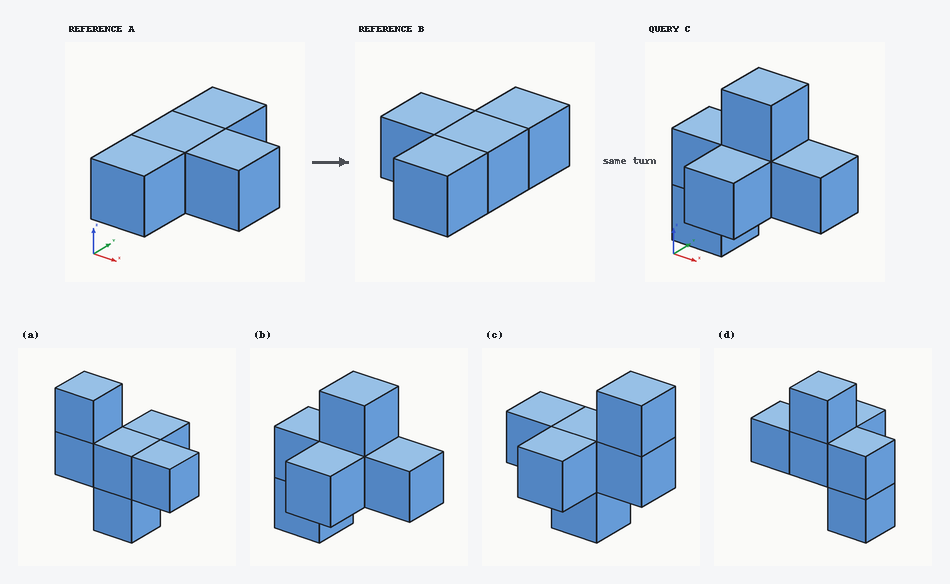}\\[0.25em]
        {\footnotesize
        \textbf{$T_1$:} Rotate $+90^\circ$ about the world $+Z$ axis.\\[-0.25em]
        $\downarrow$\\[-0.25em]
        }
        \includegraphics[width=0.31\linewidth]{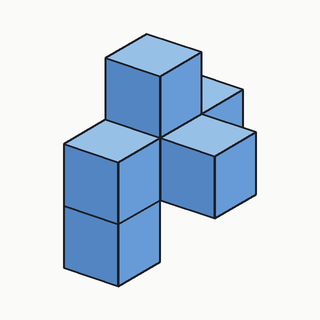}\\[-0.15em]
        {\footnotesize \textbf{$I_1$}\\[-0.1em]
        \textbf{$T_2$:} Continue another $+90^\circ$ about $+Z$.\\[-0.25em]
        $\downarrow$\\[-0.25em]
        }
        \includegraphics[width=0.31\linewidth]{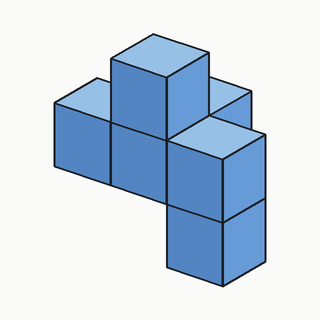}\\[-0.15em]
        {\footnotesize \textbf{$I_2$}\\[-0.1em]
        Compare with the options. \textbf{Answer: (d)}}
    \end{minipage} 
    \caption{Representative samples from \lanterntwo and \lanternthree. Text denotes mental rotation actions and images denote their resulting visual states. The 2D example stores the final state after two $90^\circ$ rotations, while the 3D example explicitly interleaves each $90^\circ$ rotation with its corresponding visual state.}
    \label{fig:lantern-data-samples}
\end{figure}

Figure \ref{fig:lantern-data-samples} shows representative training examples from \lanterntwo and \lanternthree. Each example contains a question image, textual mental actions, visual states produced by those actions, and a final textual answer. \lanterntwo stores one visual state after the complete rotation sequence, whereas \lanternthree stores a visual state after each atomic rotation.

\subsection{Comparison between Different VLMs}
We test different VLMs with different reasoning types on 2D and 3D mental rotation tasks, as well as on VSP-Nav. The results are shown in Table \ref{tab:2d-rot-on-vlms}. We observe that stronger VLMs, i.e., with strong textual reasoning capabilities, can first convert the 2D or 3D shape into coordinates. They perform textual reasoning to rotate these coordinates and get the answer. Although they can answer correctly in some cases, these models usually consume large amounts of tokens, making the reasoning process less efficient. This also leaves an open question: Are continuous visual embeddings better substrates than textual tokens for performing visual reasoning?

\paragraph{Case study: coordinate-based spatial reasoning.}
Figure~\ref{fig:qwen38-spatial-case} illustrates this behavior in a question-only \lanternthree example with Qwen3.8-27B's think mode enabled. The model correctly selects option B by encoding the query as cube coordinates, applying the specified $180^\circ$ rotation about the $Z$ axis, and matching the transformed structure to the candidates. Its emitted rationale also contains tentative pixel estimates and self-correction.

\begin{figure}[htbp]
    \centering
    \definecolor{qwensummary}{HTML}{405468}
    \newcommand{\qwenstagesummary}[2]{%
        {\normalfont\sffamily\bfseries\fontsize{9}{11}\selectfont
        \color{qwensummary}#1.\enspace#2\par}%
    }
    \tcbset{qwenoutput/.style={
        enhanced,
        colback=black!4,
        frame hidden,
        boxrule=0pt,
        sharp corners,
        borderline west={0.7pt}{0pt}{black!25},
        boxsep=0pt,
        left=8pt,right=8pt,top=6pt,bottom=6pt,
        before skip=3pt,after skip=9pt,
        fontupper=\normalfont\fontfamily{cmtt}\fontsize{9}{10.5}\selectfont,
        before upper={%
            \raggedright
            \setlength{\parindent}{0pt}%
            \setlength{\parskip}{3pt}%
            \hyphenpenalty=10000
            \exhyphenpenalty=10000
        }
    }}
    \begin{minipage}{\linewidth}
    {\normalfont\sffamily\bfseries\fontsize{10}{12}\selectfont Qwen3.8-27B: spatial reasoning case study\par}
    {\normalfont\sffamily\fontsize{9}{11}\selectfont Question only \enspace|\enspace Think mode on \enspace|\enspace Correct answer: B\par}\medskip
\qwenstagesummary{1}{Ground image directions in world coordinates}
\begin{tcolorbox}[qwenoutput]
Let me define: in isometric view, the three visible directions: right-down = +X, right-up = +Y, up = +Z. This is a standard isometric projection where X is toward the viewer\textquotesingle{}s front-right, Y is toward the back-right... Actually, let me think: the red X arrow points to the lower-right, the green Y arrow points to the upper-right, and the blue Z points up.
\end{tcolorbox}

\qwenstagesummary{2}{Switch to the analytical rotation}
\begin{tcolorbox}[qwenoutput]
This is getting too speculative. Let me think about the transformation instead: rotate +180\textdegree{} about the Z axis. This maps (x, y, z) -\textgreater{} (-x, -y, z). So the shape is flipped in the horizontal plane, and the height is preserved.

So the resulting shape should have the same height profile (z coordinates unchanged), but the x and y coordinates negated.
\end{tcolorbox}

\qwenstagesummary{3}{Encode the query as cube coordinates}
\begin{tcolorbox}[qwenoutput]
Query C: a vertical column of 3 cubes, with one cube attached to the right at the middle height. Let me define: column at (0,0,0), (0,0,1), (0,0,2); attached cube at (1,0,1)?
\end{tcolorbox}

\qwenstagesummary{4}{Infer the attachment direction}
\begin{tcolorbox}[qwenoutput]
Now the horizontal direction of the attachment: on screen, the attached cube is to the right of the column. In isometric, +X is lower-right, +Y is upper-right. The attached cube appears to the right and slightly down? Its top face center is at (810,160) or so, and the column\textquotesingle{}s middle cube\textquotesingle{}s top face would be at (775,115)? Hmm, the column\textquotesingle{}s top cube\textquotesingle{}s top face is at (775,80). The middle cube\textquotesingle{}s top face is hidden. The attached cube\textquotesingle{}s top face is at (810,160)? That\textquotesingle{}s to the lower-right of where the middle top face would be (775,115)? Lower-right = +X. So the attached cube is at (1,0,1) relative to the column at (0,0,*).
\end{tcolorbox}

\qwenstagesummary{5}{Rotate the coordinates and project back}
\begin{tcolorbox}[qwenoutput]
After 180\textdegree{} rotation about Z: (x,y,z)-\textgreater{}(-x,-y,z). The column stays at (0,0,*), the attached cube goes to (-1,0,1). On screen, -X is upper-left. So the attached cube should appear to the upper-left of the column at middle level.
\end{tcolorbox}

\qwenstagesummary{6}{Match and eliminate answer candidates}
\begin{tcolorbox}[qwenoutput]
In option b, the attached cube relative to the column\textquotesingle{}s middle cube: the attached cube\textquotesingle{}s top face appears to the left and slightly up? The column\textquotesingle{}s middle cube\textquotesingle{}s top face is hidden; the column\textquotesingle{}s top cube\textquotesingle{}s top face is at (395,400). The middle cube\textquotesingle{}s top face would be at (395,445) (one level down \textasciitilde{}45). The attached cube\textquotesingle{}s top face is at (345,435)? That\textquotesingle{}s left and slightly up relative to (395,445): left-up = -X. Yes! So option b shows the attached cube at -X, which matches the 180\textdegree{} rotation of C if the original attachment was at +X.

But let me double-check the options: option a is identical to C (protrusion to the right). Option b is protrusion to the left. Option c and d are flat shapes (different), which would correspond to rotations about other axes (e.g., 90\textdegree{} about X or Y). So the answer is b.

Therefore: \textless{}answer\textgreater{}B\textless{}/answer\textgreater{}.
\end{tcolorbox}

    \end{minipage}
    \caption{Sampled reasoning trajectory from Qwen3.8-27B. We randomly select a sample from \lanternthree test set and only input the question to the model. Blue-gray headings summarize the reasoning stages; shaded areas shows the original output from the model. The model constructs a sparse coordinate representation, applies $(x,y,z)\mapsto(-x,-y,z)$, and checks the answer candidates. The model answers correctly, but its reasoning is long and repetitive.}
    \label{fig:qwen38-spatial-case}
\end{figure}

\begin{table}[htbp]
    \centering
    \caption{Evaluation results of VLMs on the test sets of \lanterntwo, \lanternthree, and VSP-Nav under different reasoning-input settings. Think Mode is the model's built-in capability from their pre-training. Qwen2.5-VL-7B-Instruct and Qwen3-VL-8B-Instruct do not support think mode by default. Qwen3.5-9B, Qwen3.6-27B, and Qwen3.8-27B support think mode. We test these models in both modes. \ding{55} indicates the think mode is disabled; \ding{51} indicates the think mode is enabled. Q denotes the question input, RT denotes reasoning text, and I denotes intermediate reasoning images. When testing with the Q + RT setting, we replace the intermediate reasoning images with token "[IMAGINATION]".}
    \label{tab:2d-rot-on-vlms}
    \small
    \setlength{\tabcolsep}{4pt}
    \renewcommand{\arraystretch}{1.08}
    \resizebox{0.9\linewidth}{!}{%
    \begin{tabular}{@{}lclrrrrrr@{}}
        \toprule
        \multirow[c]{2}{*}{Model} & \multirow[c]{2}{*}{Think Mode} & \multirow[c]{2}{*}{Eval Setting} & \multicolumn{2}{c}{\lanterntwo} & \multicolumn{3}{c}{\lanternthree} & \multirow[c]{2}{*}{VSP-Nav} \\
        \cmidrule(lr){4-5} \cmidrule(lr){6-8}
        & & & \multicolumn{1}{c}{ID} & \multicolumn{1}{c}{OOD} & \multicolumn{1}{c}{SC-ID} & \multicolumn{1}{c}{C-ID} & \multicolumn{1}{c}{OOD} & \\
        \midrule
        \multirow[c]{3}{*}{Qwen2.5-VL-7B-Instruct} & \multirow[c]{3}{*}{\ding{55}} & Question only & 22.00 & 23.80 & 21.25 & 23.00 & 21.40 & 6.50 \\
                      &   & Q + RT & 23.50 & 23.20 & 21.00 & 20.25 & 21.00 & 21.50 \\
                      &   & Q + RT + I & 26.25 & 26.20 & 58.25 & 60.50 & 59.60 & 24.83 \\
        \midrule
        \multirow[c]{3}{*}{Qwen3-VL-8B-Instruct} & \multirow[c]{3}{*}{\ding{55}} & Question only & 23.00 & 21.80 & 9.75 & 9.00 & 9.20 & 1.33 \\
                    &   & Q + RT & 20.50 & 21.00 & 8.25 & 7.00 & 7.40 & 0.67 \\
                    &   & Q + RT + I & 33.75 & 38.40 & 51.50 & 53.00 & 46.60 & 0.50 \\
        \midrule
        \multirow[c]{6}{*}[-\dimexpr(\aboverulesep+\belowrulesep+\cmidrulewidth)/2\relax]{Qwen3.5-9B} & \multirow[c]{3}{*}{\ding{55}} & Question only & 29.25 & 24.20 & 18.75 & 18.50 & 19.20 & 0.17 \\
                   &   & Q + RT & 27.25 & 23.00 & 23.50 & 23.25 & 21.40 & 3.00 \\
                   &   & Q + RT + I & 30.50 & 29.80 & 59.00 & 58.50 & 61.80 & 42.50 \\
        \cmidrule(l){2-9}
                   & \multirow[c]{3}{*}{\ding{51}} & Question only & 42.50 & 41.80 & 26.00 & 22.25 & 23.80 & 61.00 \\
                   &   & Q + RT & 43.75 & 40.00 & 25.50 & 24.00 & 28.00 & 81.00 \\
                   &   & Q + RT + I & 44.50 & 41.20 & 45.25 & 40.50 & 46.80 & 75.67 \\
        \midrule
        \multirow[c]{6}{*}[-\dimexpr(\aboverulesep+\belowrulesep+\cmidrulewidth)/2\relax]{Qwen3.6-27B} & \multirow[c]{3}{*}{\ding{55}} & Question only & 43.50 & 45.60 & 22.25 & 18.25 & 21.60 & 0.00 \\
                    &   & Q + RT & 48.25 & 46.20 & 13.50 & 12.00 & 12.20 & 15.33 \\
                    &   & Q + RT + I & 28.50 & 31.20 & 68.25 & 66.00 & 68.40 & 2.83 \\
        \cmidrule(l){2-9}
                    & \multirow[c]{3}{*}{\ding{51}} & Question only & 56.00 & 47.80 & 26.50 & 25.25 & 26.60 & 71.17 \\
                    &   & Q + RT & 60.50 & 46.00 & 30.50 & 26.00 & 21.80 & 91.83 \\
                    &   & Q + RT + I & 55.50 & 51.60 & 68.25 & 63.75 & 66.00 & 95.33 \\
        \midrule
        \multirow[c]{6}{*}[-\dimexpr(\aboverulesep+\belowrulesep+\cmidrulewidth)/2\relax]{Qwen3.8-27B} & \multirow[c]{3}{*}{\ding{55}} & Question only & 67.00 & 64.20 & 26.25 & 29.25 & 26.40 & 16.00 \\
                    &   & Q + RT & 65.00 & 63.20 & 29.25 & 27.50 & 27.40 & 48.00 \\
                    &   & Q + RT + I & 61.25 & 60.60 & 53.75 & 53.75 & 55.00 & 95.67 \\
        \cmidrule(l){2-9}
                    & \multirow[c]{3}{*}{\ding{51}} & Question only & 93.50 & 85.40 & 39.75 & 39.00 & 32.20 & 88.00 \\
                    &   & Q + RT & 93.00 & 88.60 & 32.75 & 32.00 & 32.40 & 99.67 \\
                    &   & Q + RT + I & 87.50 & 84.00 & 88.75 & 85.75 & 87.80 & 99.83 \\
        \bottomrule
    \end{tabular}%
    }
\end{table}

\section{Additional Experiment Results}

\begin{table}[htbp]
    \centering
    \caption{Maze-navigation results across different training datasets and grid sizes. Accuracy and per-grid-size results are reported in percent.}
    \label{tab:additional-navigation-results}
    \setlength{\tabcolsep}{4pt}
    \renewcommand{\arraystretch}{1.08}
    \resizebox{\textwidth}{!}{%
    \begin{tabular}{@{}llcrrrrrrr@{}}
        \toprule
        Training Data & Method & WM Objective & Accuracy & $3{\times}3$ & $4{\times}4$ & $5{\times}5$ & $6{\times}6$ & $7{\times}7$ & $8{\times}8$ \\
        \midrule
        \shortstack[l]{Proprietary interleaved data \\ $+$ \thinkmorphnavorigin} & \shortstack[l]{ThinkMorph \\ (BAGEL)} & \ding{51} & 82.50 & 93 & 95 & 94 & 78 & 74 & 61 \\
        \midrule
        / & Qwen2.5-VL-7B-Inst. & \ding{55} & 6.50 & 23 & 4 & 5 & 3 & 2 & 2 \\
        \midrule
        \multirow{4}{*}{\thinkmorphnavorigin}
            & Qwen2.5-VL-7B-Inst. SFT & \ding{55} & 43.17 & 79 & 65 & 50 & 34 & 17 & 14 \\
            & LatentUM & \ding{51} & 52.00 & 90 & 75 & 59 & 42 & 24 & 22 \\
            & Mirage & \ding{51} & 42.50 & 79 & 58 & 56 & 29 & 21 & 12 \\
            & WM-VLM (Ours) & \ding{51} & 50.00 & 98 & 83 & 63 & 31 & 18 & 7 \\
        \bottomrule
    \end{tabular}%
    }
\end{table}

\begin{figure}[htbp]
    \centering
    \includegraphics[width=0.95\textwidth]{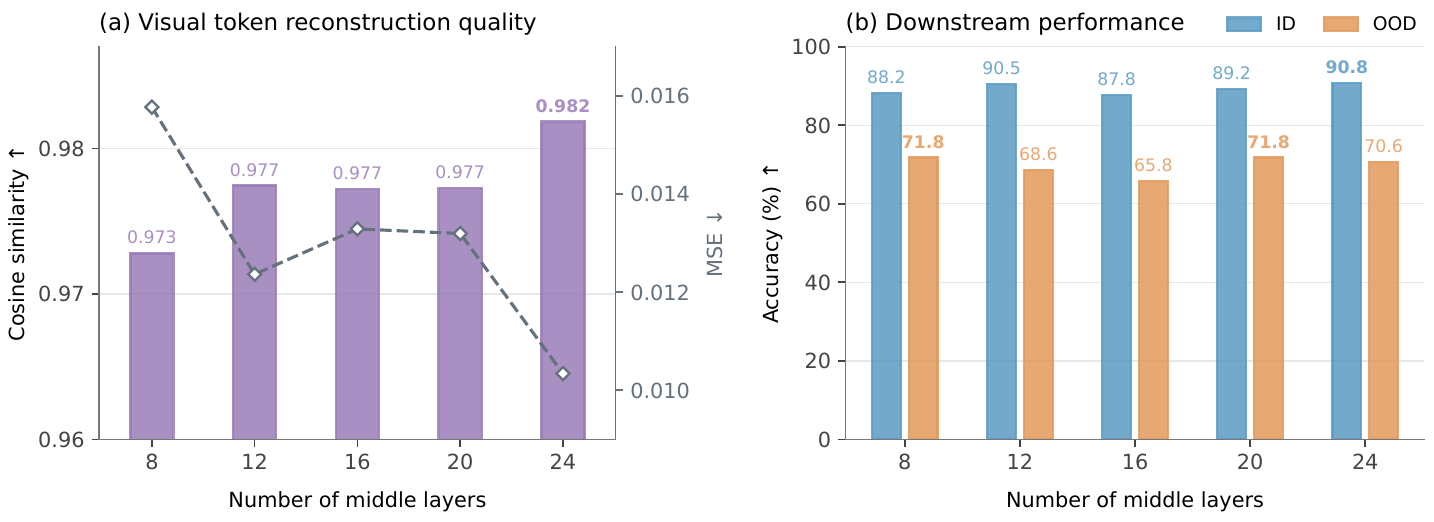}
    \caption{Effect of the number of middle layers on model performance.}
    \label{fig:middle-layer-ablation}
\end{figure}

\begin{table}[htbp]
    \centering
    \caption{Hyperparameters for the two-stage training procedure on \lanterntwo.}
    \label{tab:training-hyperparameters}
    \setlength{\tabcolsep}{8pt}
    \renewcommand{\arraystretch}{1.08}
    \resizebox{0.8\linewidth}{!}{%
    \begin{tabular}{@{}lcc@{}}
        \toprule
        \textbf{Hyperparameter} & \textbf{Stage 1} & \textbf{Stage 2} \\
        \midrule
        Initialization & Qwen2.5-VL-7B-Instruct & Stage 1 checkpoint \\
        \midrule
        Learning rate & 1e-4 & 1e-5 \\
        Optimizer & AdamW & AdamW \\
        LR scheduler & Constant & Cosine \\
        Warmup ratio & 0 & 0.03 \\
        Weight decay & 0 & 0.01 \\
        Per-device batch size & 1 & 1 \\
        GPU Model & H200 & H200 \\
        Number of GPUs & 8 & 8 \\
        Gradient accumulation steps & 1 & 1 \\
        Effective global batch size & 8 & 8 \\
        Precision & BF16 & BF16 \\
        Flow/MSE loss weight & 1 & 0 \\
        Pixel loss weight & 0 & 0 \\
        CE loss weight & 0 & 1 \\
        Loss formula & 1.0 $\times$ flow MSE & 1.0 $\times$ CE \\
        VLM backbone & Frozen & Trainable \\
        Vision encoder & Frozen & Frozen \\
        Latent-generation branch & Trainable & Trainable \\
        \midrule
        Trainable parameters & 0.959B & 8.572B \\
        Frozen parameters & 8.289B & 0.677B \\
        Total parameters & 9.248B & 9.248B \\
        Trainable ratio & 10.37\% & 92.68\% \\
        \bottomrule
    \end{tabular}%
    }
\end{table}

\begin{figure}[htbp]
    \centering
    \includegraphics[width=0.8\textwidth]{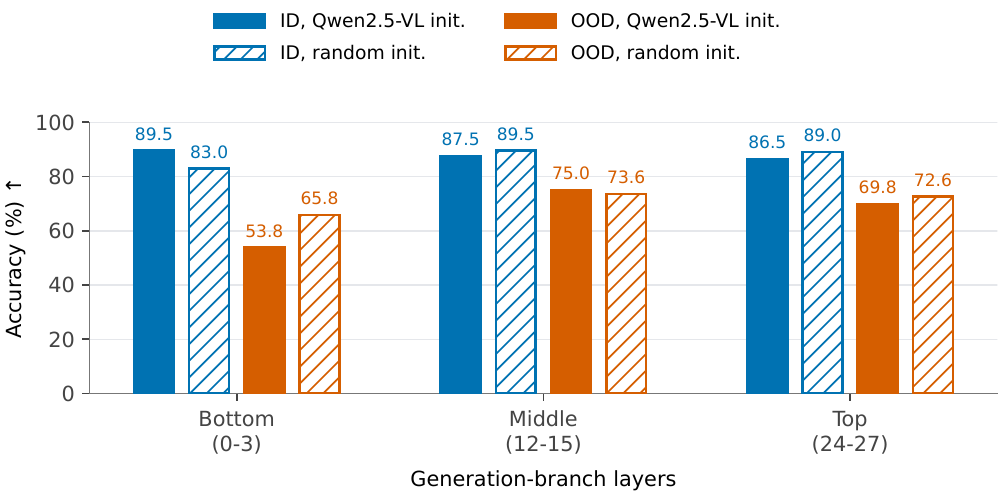}
    \caption{Effect of generation-branch initialization on downstream performance across different insertion depths.}
    \label{fig:generation-branch-initialization}
\end{figure}

\paragraph{Initialization is not decisive.} In our main experiments, generation branch is initialized with the same weights as the understanding branch. To weigh the factor of the intialization method, we conduct an ablation study by initializing the generation branch with random weights. As shown in Figure \ref{fig:generation-branch-initialization}, in all three layer placement settings, neither initialization method shows a clear advantage over the other. This indicates that the initialization method is not decisive for the final performance.

\paragraph{Experiments on visual planning tasks.} We further evaluate our model on visual planning tasks, which require predicting a sequence of actions to reach a goal state. We train on \thinkmorphnavorigin and evaluate on VSP-Nav. As shown in Table \ref{tab:additional-navigation-results}, our model, which generates continuous visual tokens as intermediate thoughts, shows competitive performance compared with baselines. BAGEL generates images as intermediate thoughts and achieves the best performance, likely because it was pre-trained on large-scale interleaved data. In contrast, on the Tetris tasks, where BAGEL appears not to benefit from similar pre-training data, it underperforms our model and even collapses during training.

\paragraph{Additional results on generation layers.} We further vary the number of generation layers from 8 to 24. As shown in Figure \ref{fig:middle-layer-ablation}, increasing the generation-branch depth generally improves visual-state reconstruction quality, but does not consistently improve downstream accuracy. This suggests that better reconstruction alone does not necessarily translate into better reasoning performance.
 
\begin{figure}[p]
    \centering
    \includegraphics[width=0.8\linewidth]{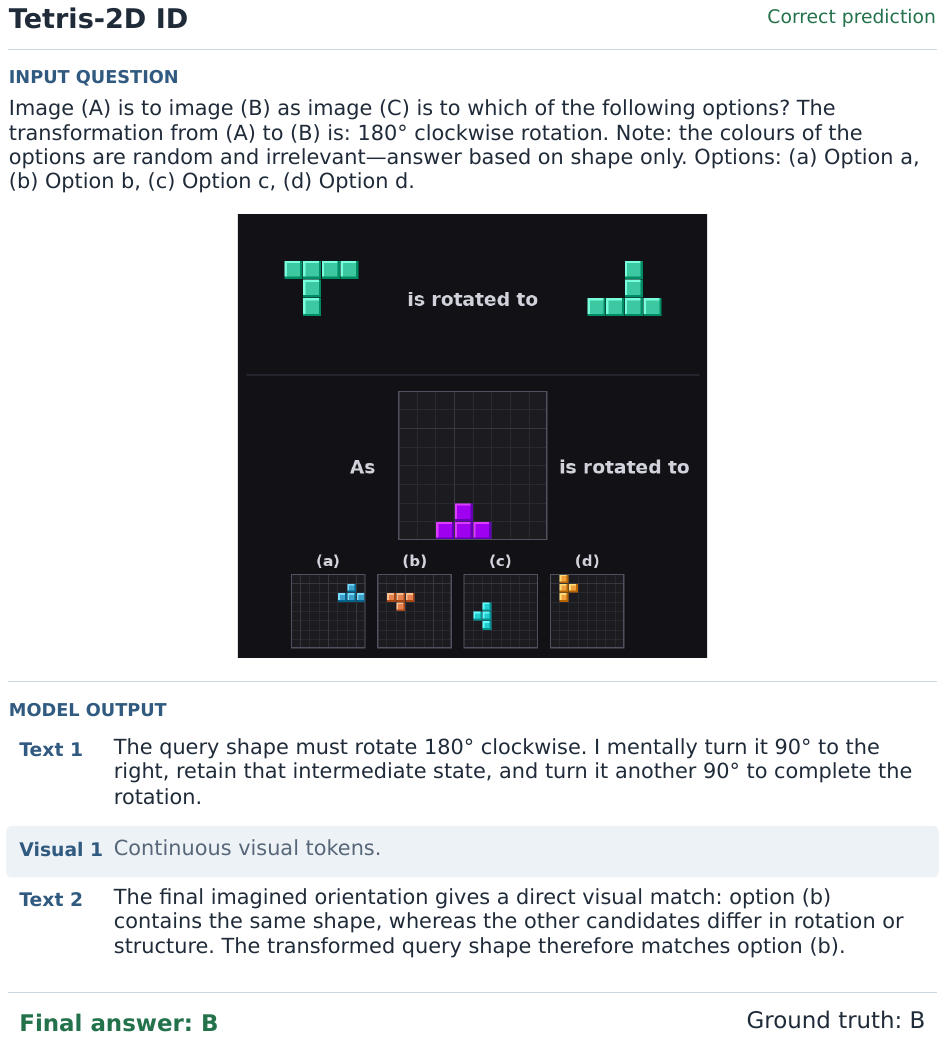}
    \caption{Correct \modelname prediction on \lanterntwo-ID (sample 17). The question specifies a $180^\circ$ clockwise rotation. The model predicts option B, matching the ground-truth answer.}
    \label{fig:model-output-2d-id}
\end{figure}

\begin{figure}[p]
    \centering
    \includegraphics[width=0.8\linewidth]{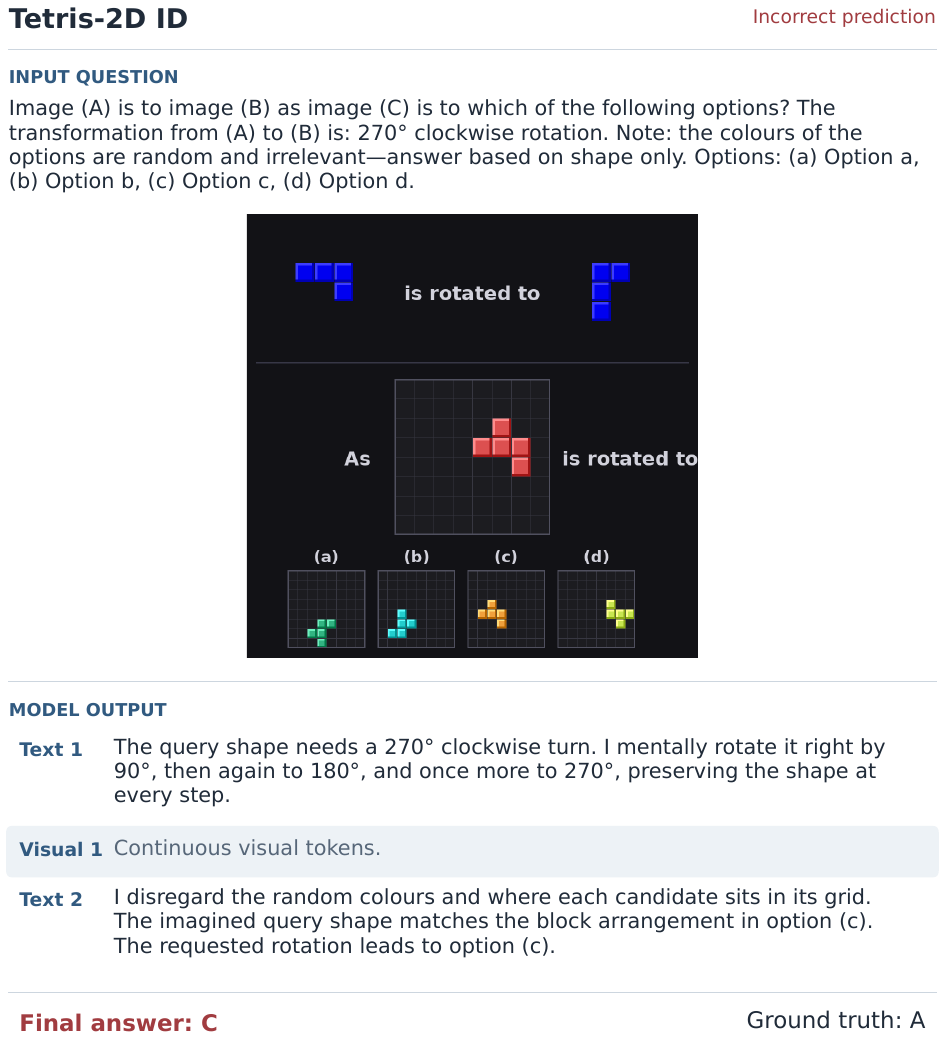}
    \caption{Incorrect \modelname prediction on \lanterntwo-ID (sample 48). The question specifies a $270^\circ$ clockwise rotation. The model predicts option C, whereas the ground-truth answer is A.}
    \label{fig:model-output-2d-id-wrong}
\end{figure}

\begin{figure}[p]
    \centering
    \includegraphics[width=0.8\linewidth]{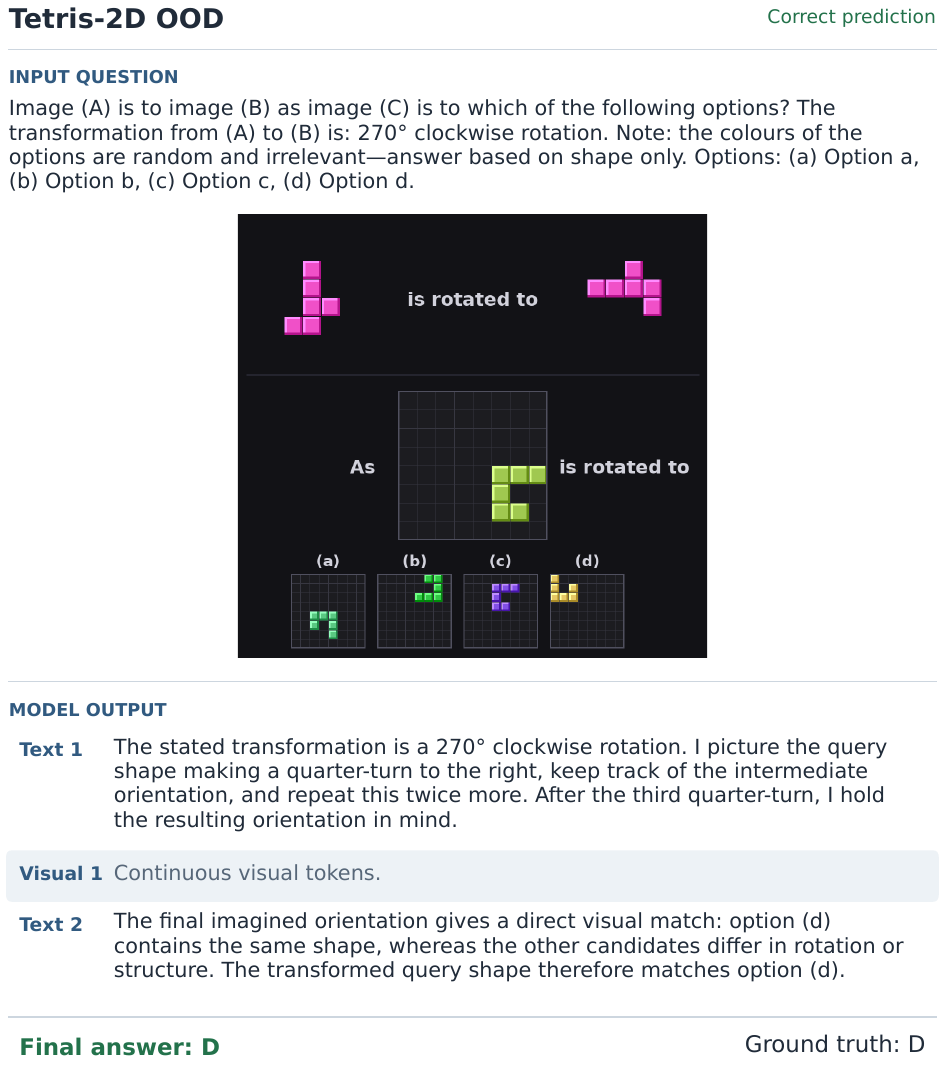}
    \caption{Correct \modelname prediction on \lanterntwo-OOD (sample 11). The question specifies a $270^\circ$ clockwise rotation. The model predicts option D, matching the ground-truth answer.}
    \label{fig:model-output-2d-ood}
\end{figure}

\begin{figure}[p]
    \centering
    \includegraphics[width=0.8\linewidth]{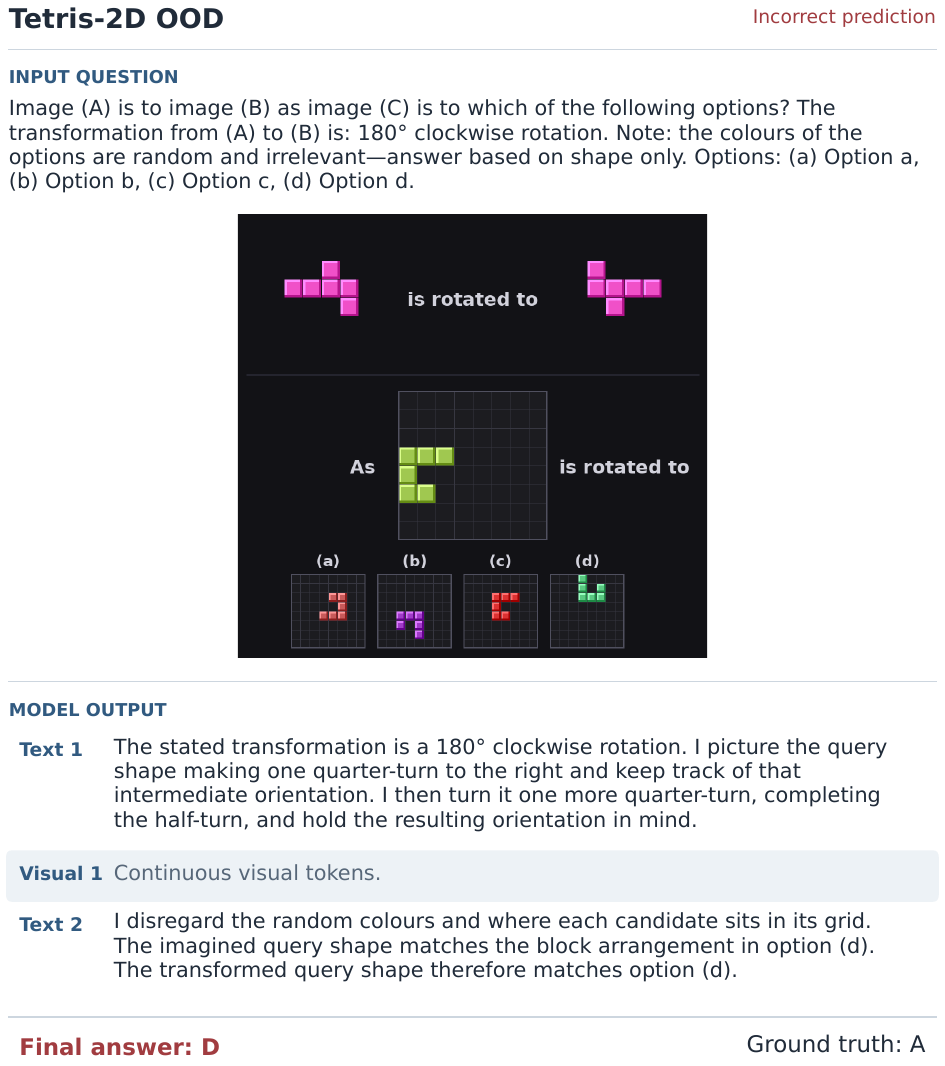}
    \caption{Incorrect \modelname prediction on \lanterntwo-OOD (sample 48). The question specifies a $180^\circ$ clockwise rotation. The model predicts option D, whereas the ground-truth answer is A.}
    \label{fig:model-output-2d-ood-wrong}
\end{figure}

\begin{figure}[p]
    \centering
    \includegraphics[width=0.8\linewidth]{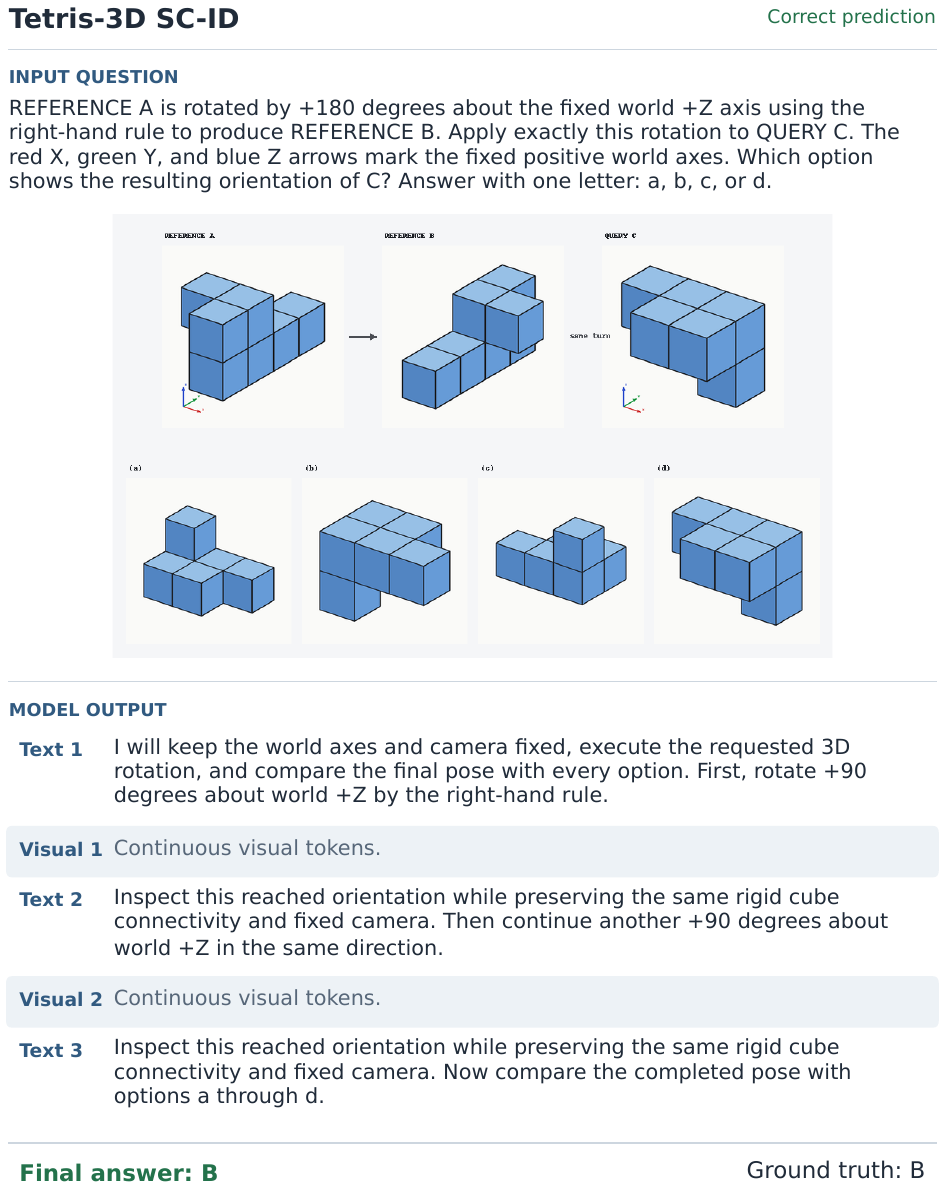}
    \caption{Correct \modelname prediction on \lanternthree-SC-ID (sample 81). The question specifies a $+180^\circ$ rotation about the fixed world $+Z$ axis. The model describes two successive $+90^\circ$ turns and predicts option B, matching the ground-truth answer.}
    \label{fig:model-output-3d-sc-id}
\end{figure}

\begin{figure}[p]
    \centering
    \includegraphics[width=0.8\linewidth]{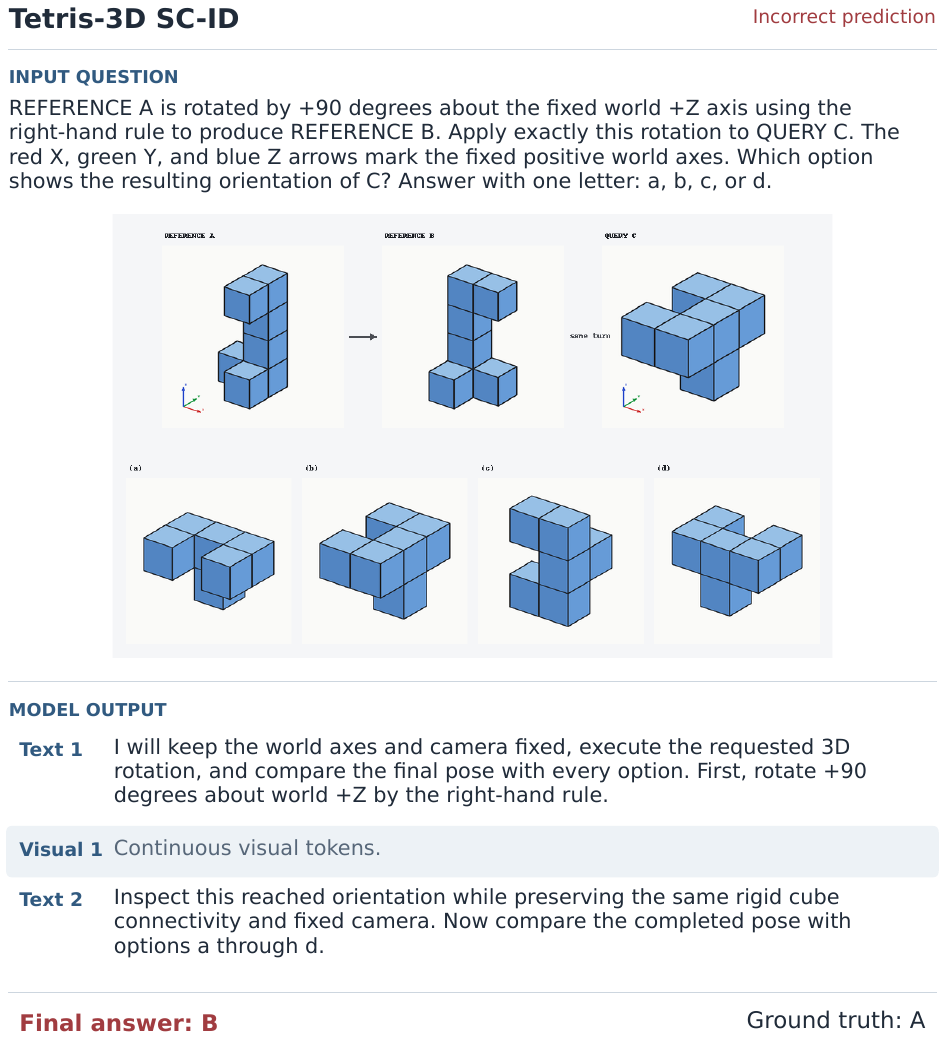}
    \caption{Incorrect \modelname prediction on \lanternthree-SC-ID (sample 152). The question specifies a $+90^\circ$ rotation about the fixed world $+Z$ axis. The model predicts option B, whereas the ground-truth answer is A.}
    \label{fig:model-output-3d-sc-id-wrong}
\end{figure}

\begin{figure}[p]
    \centering
    \includegraphics[width=0.8\linewidth]{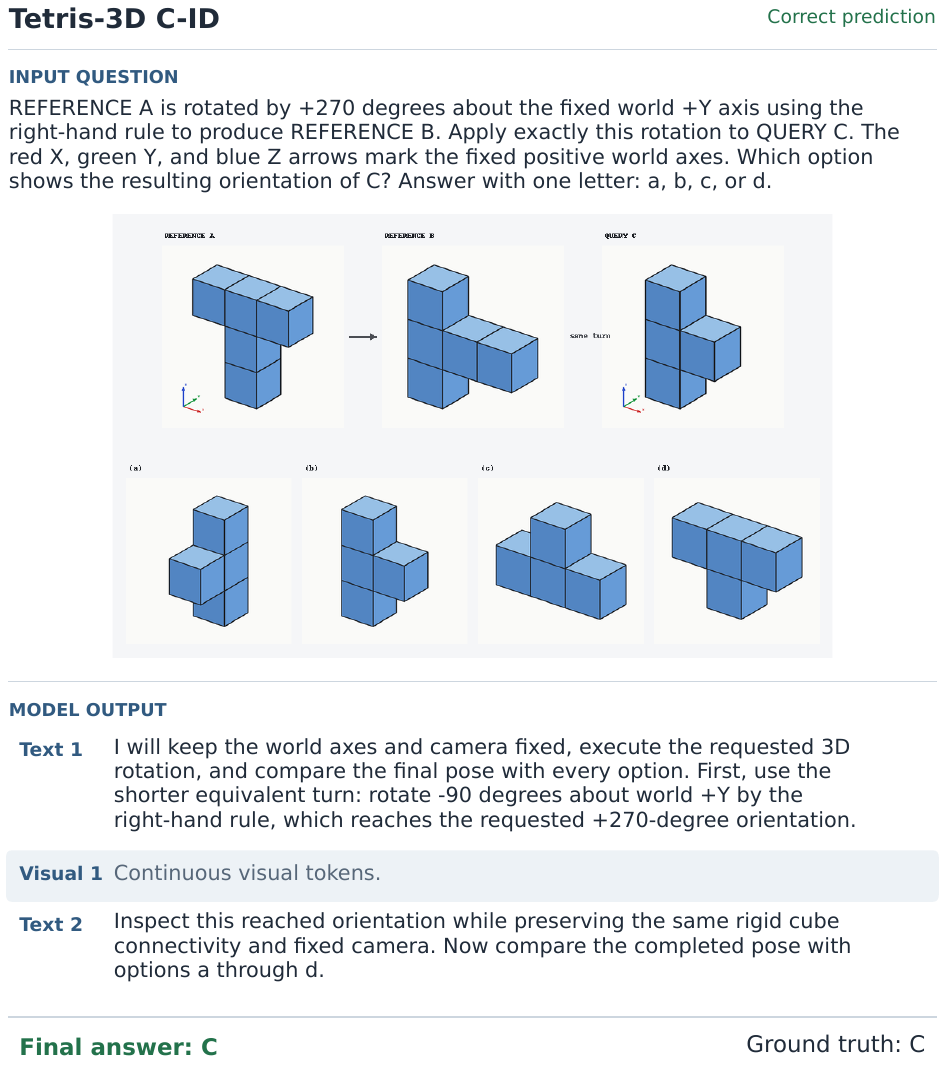}
    \caption{Correct \modelname prediction on \lanternthree-C-ID (sample 34). The question specifies a $+270^\circ$ rotation about the fixed world $+Y$ axis. The model describes the equivalent $-90^\circ$ turn and predicts option C, matching the ground-truth answer.}
    \label{fig:model-output-3d-c-id}
\end{figure}

\begin{figure}[p]
    \centering
    \includegraphics[width=0.8\linewidth]{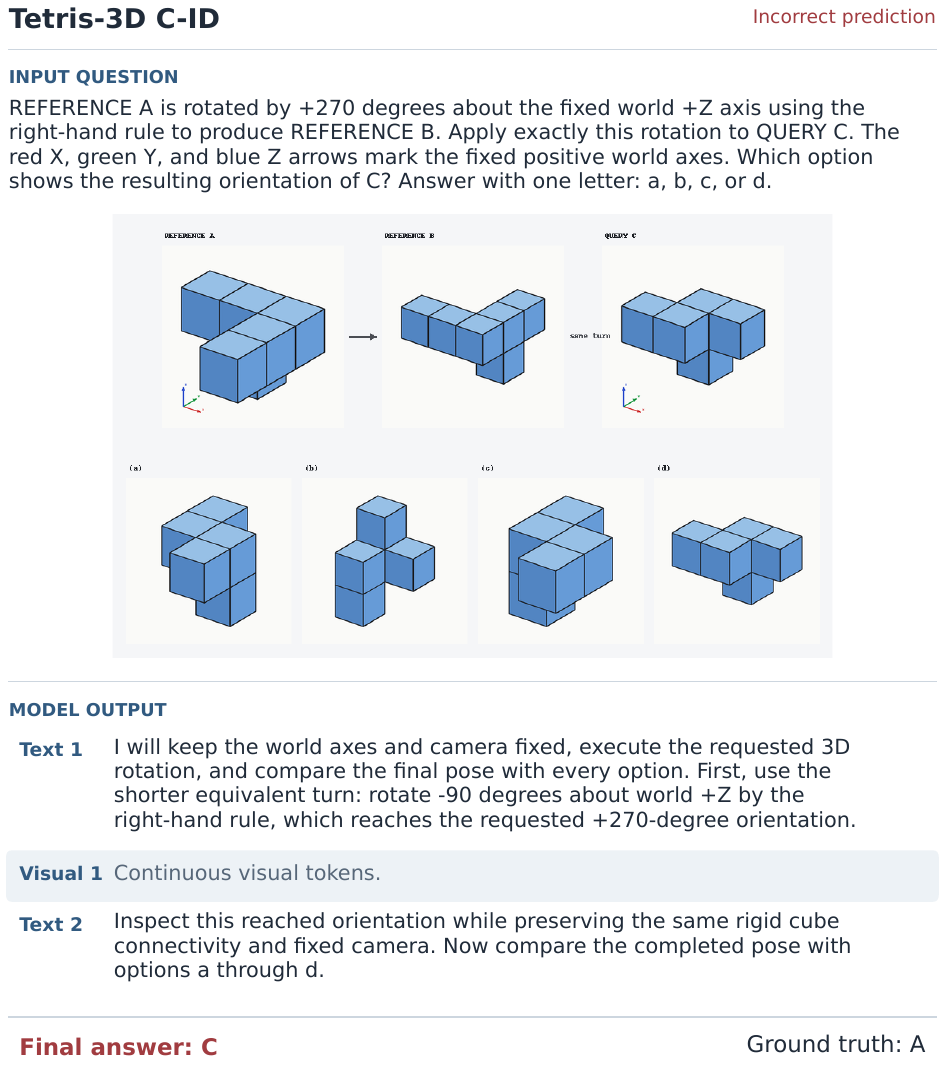}
    \caption{Incorrect \modelname prediction on \lanternthree-C-ID (sample 40). The question specifies a $+270^\circ$ rotation about the fixed world $+Z$ axis. The model describes the equivalent $-90^\circ$ turn and predicts option C, whereas the ground-truth answer is A.}
    \label{fig:model-output-3d-c-id-wrong}
\end{figure}

\begin{figure}[p]
    \centering
    \includegraphics[width=0.8\linewidth]{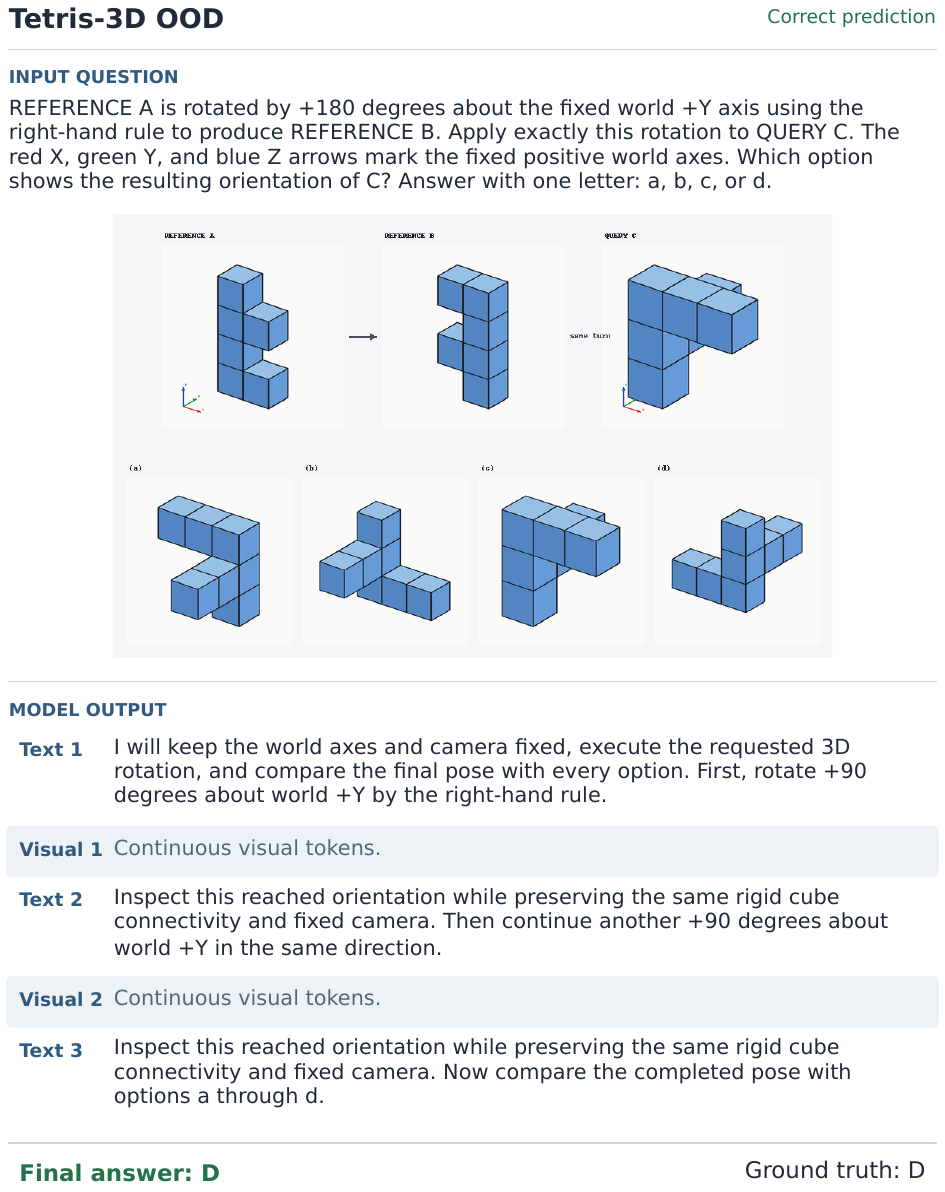}
    \caption{Correct \modelname prediction on \lanternthree-OOD (sample 19). The question specifies a $+180^\circ$ rotation about the fixed world $+Y$ axis. The model describes two successive $+90^\circ$ turns and predicts option D, matching the ground-truth answer.}
    \label{fig:model-output-3d-ood}
\end{figure}

\begin{figure}[p]
    \centering
    \includegraphics[width=0.8\linewidth]{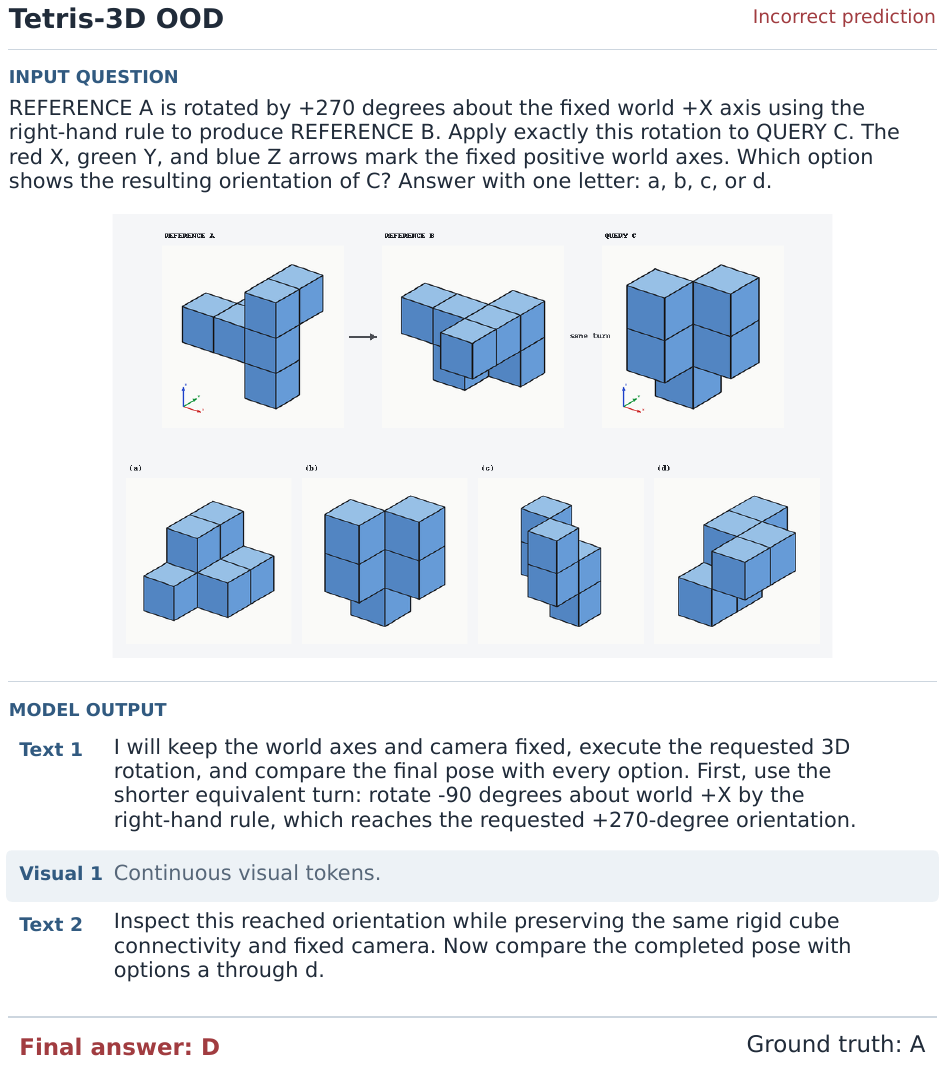}
    \caption{Incorrect \modelname prediction on \lanternthree-OOD (sample 24). The question specifies a $+270^\circ$ rotation about the fixed world $+X$ axis. The model describes the equivalent $-90^\circ$ turn and predicts option D, whereas the ground-truth answer is A.}
    \label{fig:model-output-3d-ood-wrong}
\end{figure}

\paragraph{Model Output Examples.}
Figures~\ref{fig:model-output-2d-id}--\ref{fig:model-output-3d-ood-wrong} show one correct and one incorrect prediction from each of the five Tetris test splits. Each example includes the input question and image, textual reasoning, prediction, and ground-truth answer. Shaded rows mark intermediate continuous visual tokens.

\clearpage

\section{Future Work}
Our training paradigm also supports pretraining a VLM with an internal world model, provided that an interleaved visual-textual reasoning dataset is available. However, existing interleaved datasets are primarily derived from static sources, such as webpages. Ideally, such a dataset would instead capture embodied experience, in which every physical action is paired with its resulting visual observation. Complete trajectories, including the initial instruction and observation, interleaved actions and observations, and final outcome, could be collected from either real-world robots or simulated embodied agents. A VLM pretrained on these trajectories could learn to mentally simulate the consequences of actions, potentially improving spatial reasoning and agentic planning. We leave the construction of such a dataset and the exploration of this training paradigm to future work.

\end{document}